\documentclass[acmtog]{acmart}

\AtBeginDocument{%
  \providecommand\BibTeX{{%
    \normalfont B\kern-0.5em{\scshape i\kern-0.25em b}\kern-0.8em\TeX}}}

\copyrightyear{2025}
\acmYear{2025}
\setcopyright{acmlicensed}\acmConference[SIGGRAPH '25 Conference
Proceedings]{Special Interest Group on Computer Graphics and
Interactive Techniques Conference Conference Proceedings}{August
10--14, 2025}{Vancouver, BC, Canada}
\acmBooktitle{Special Interest Group on Computer Graphics and
Interactive Techniques Conference Conference Proceedings (SIGGRAPH '25
Conference Proceedings), August 10--14, 2025, Vancouver, BC, Canada}
\acmDOI{10.1145/3721238.3730639}
\acmISBN{979-8-4007-1540-2/2025/08}

\acmDOI{10.1145/3721238.3730639}

\usepackage[boxed,ruled,linesnumbered]{algorithm2e} 

\SetCommentSty{mycommfont}
\let\oldnl\nl
\newcommand{\nonl}{\renewcommand{\nl}{\let\nl\oldnl}}

\SetAlFnt{\small}
\SetAlCapFnt{\small}
\SetAlCapNameFnt{\small}
\SetAlCapHSkip{0pt}
\IncMargin{-\parindent}

\newlength\savedwidth

\usepackage{booktabs} 
\usepackage{enumitem}
\usepackage{amsmath}
\usepackage{color, colortbl}
\definecolor{lightgray}{gray}{0.9}
\usepackage{xcolor}
\usepackage{wrapfig}
\DeclareMathOperator*{\argmax}{\arg\!\max}
\DeclareMathOperator*{\argmin}{\arg\!\min}
\usepackage{soul}
\usepackage{multirow}
\theoremstyle{assumption}

\theoremstyle{definition}

\newcommand{\mainpaper}[1]{\textcolor{blue}{#1}}

\newcommand{\Samples}{\mathcal{S}}
\newcommand{\Samplesprior}{\mathcal{S}^P}
\newcommand{\Samplesrealtime}{\mathcal{S}^t}
\newcommand{\sample}{\mathbf{s}}
\newcommand{\sampleprior}{\mathbf{s}^p}
\newcommand{\samplerealtime}{\mathbf{s}^t}

\newcommand{\Views}{\mathcal{V}}
\newcommand{\Viewsobserve}{\mathcal{V}^o}
\newcommand{\Viewsvisited}{\mathcal{V}^v}
\newcommand{\Viewsunvisited}{\mathcal{V}^{uv}}
\newcommand{\Viewsprior}{\mathcal{V}^P}
\newcommand{\Viewsrealtime}{\mathcal{V}^R}

\newcommand{\view}{\mathbf{v}}
\newcommand{\viewdir}{\mathbf{o}}
\newcommand{\viewprior}{\mathbf{v}^p}

\newcommand{\prob}{p}
\newcommand{\probprior}{p^p}

\newcommand{\probwrap}{q}

\newcommand{\cov}{c}
\newcommand{\vis}{\delta}
\newcommand{\target}{\mathcal{T}}
\newcommand{\targetprior}{\mathcal{T}^p}
\newcommand{\targetrealtime}{\mathcal{T}^t}

\newcommand{\image}{\mathcal{I}}
\newcommand{\diffarea}{\mathcal{I}^{\text{diff}}}
\newcommand{\pointcloud}{\mathcal{P}}

\newcommand{\nextview}{\textit{next\_view}~}
\newcommand{\prevview}{\textit{prev\_view}~}
\newcommand{\priorviews}{\textit{prior\_views}}
\newcommand{\candidateview}{\textit{candidate\_views}~}

\usepackage{amsmath}

\newcommand{\eg}{\textit{e.g., }}

\usepackage{bbding}

\begin{document}
\title{Aerial Path Online Planning for Urban Scene Updation}

\author{Mingfeng Tang}
\authornote{Both authors contributed equally to this research.}
\affiliation{%
  \institution{Shenzhen University}
  \city{Shenzhen}
  \country{China}
}
\email{t20010510@gmail.com}

\author{Ningna Wang}
\affiliation{%
  \institution{University of Texas at Dallas}
  \state{Texas}
  \country{USA}
}
\authornotemark[1]
\email{ningna.wang@utdallas.edu}

\author{Ziyuan Xie}
\affiliation{%
  \institution{Shenzhen University}
  \city{Shenzhen}
  \country{China}
}
\email{turneraaaaa@gmail.com}

\author{Jianwei Hu}
\affiliation{%
  \institution{QiYuan Lab}
  \city{Beijing}
  \country{China}
}
\email{hujianwei@qiyuanlab.com}

\author{Ke Xie}
\affiliation{%
  \institution{Shenzhen University}
  \city{Shenzhen}
  \country{China}
}
\email{Ke.xie.siat@gmail.com}

\author{Xiaohu Guo}
\affiliation{
  \institution{University of Texas at Dallas}
  \state{Texas}
  \country{USA}
}
\email{xguo@utdallas.edu}

\author{Hui Huang}\authornote{Corresponding author: Hui Huang (hhzhiyan@gmail.com)}
\affiliation{%
  \institution{Shenzhen University}
  \city{Shenzhen}
  \country{China}
}
\email{hhzhiyan@gmail.com}

\renewcommand\shortauthors{M. Tang, N. Wang, Z. Xie, J. Hu, K. Xie, X. Guo, and H. Huang}

\begin{abstract}
We present the first \textit{scene-update} aerial path planning algorithm specifically designed for detecting and updating change areas in urban environments. While existing methods for large-scale 3D urban scene reconstruction focus on achieving high accuracy and completeness, they are inefficient for scenarios requiring periodic updates, as they often re-explore and reconstruct entire scenes, wasting significant time and resources on unchanged areas. To address this limitation, our method leverages prior reconstructions and change probability statistics to guide UAVs in detecting and focusing on areas likely to have changed. Our approach introduces a novel \textit{changeability heuristic} to evaluate the likelihood of scene changes, driving the planning of two flight paths: a \textit{prior path} informed by static priors and a dynamic \textit{real-time path} that adapts to newly detected changes. Extensive experiments on real-world urban datasets demonstrate that our method significantly reduces flight time and computational overhead while maintaining high-quality updates comparable to full-scene re-exploration and reconstruction. These contributions pave the way for efficient, scalable, and adaptive UAV-based scene updates in complex urban environments.
\end{abstract}




\begin{teaserfigure}
    \centering
     \includegraphics[width=\linewidth]{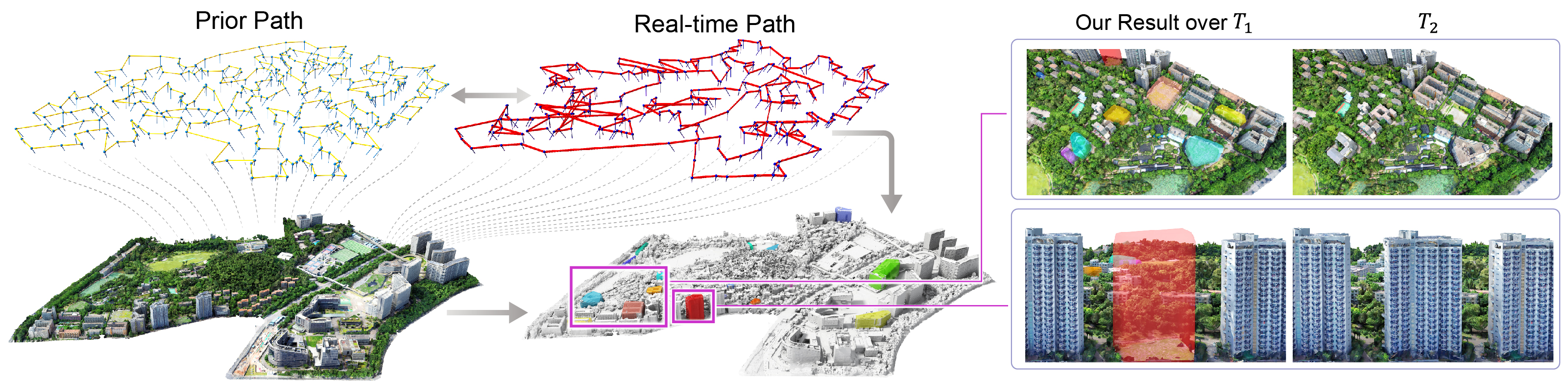}
    \caption{We present the first aerial trajectory planning method for urban scene updates. Starting with a previously explored scene ($T_1$, bottom left), our method employs a two-step approach: a \textit{prior path} based on prior probabilities and a \textit{real-time path} triggered by detected changes. The output includes convex hulls of changed areas (bottom middle), enabling accurate 3D scene updates with significantly reduced flight time.}
    \label{fig:teaser}
\end{teaserfigure}

\maketitle

\section{Introduction}
\label{sec:intro}

Aerial path planning for large-scale 3D urban scene acquisition using Unmanned Aerial Vehicles (UAVs) has garnered significant attention over recent decades~\cite{hepp2018learn, smith2018aerial, liu2021aerial, liu2022learning, zhang2021continuous, zhou2020offsite}. The primary objective of these approaches is to achieve high-accuracy 
3D reconstructions while operating within the physical constraints of UAVs, such as battery life and flight speed.

Existing aerial path planning methods are well-suited for reconstructing scenes from scratch but are inefficient for scenarios requiring periodic updates. For instance, many urban scenes are reconstructed at regular intervals—$T_1$ represents the time of a prior exploration and reconstruction, and $T_2$ represents the current revisit. Traditional methods often require re-exploring and reconstructing the entire scene, even though most parts of large-scale urban environments remain static over time. For example, in the Yingrenshi scene from UrbanBIS~\cite{yang2023urbanbis}, re-flying and reconstructing the entire scene would take over two weeks, despite the majority of the scene being unchanged. This inefficiency highlights the urgent need for a targeted approach to detect and update only the regions that have changed.

In this paper, we present the first dedicated framework for aerial path planning specifically designed for change detection in large-scale urban environments. Our method addresses the limitations of traditional approaches by enabling real-time detection and extraction of change areas in previously explored scenes. By leveraging a pre-existing $T_1$ reconstruction and change probability statistics derived from remote sensing datasets, our approach identifies and prioritizes areas likely to have changed, avoiding redundant exploration of static regions. 

Our proposed framework comprises two key components: a \textit{prior path} and a \textit{real-time path}. To plan these paths, we introduce a \textit{changeability heuristic} in Sec.~\ref{sec:changeability}, which evaluates the prioritization of views and samples based on their coverage and change detection potential. For the prior path, we utilize the $T_1$ reconstructed model and statistical data to evaluate the changeability of views and samples. 
As the drone traverses the prior path, any detected change area triggers an update to the real-time path, prioritizing the detection and extraction of the entire change area. 
Experimental results demonstrate that our online algorithm effectively detects and extracts change areas, enabling previously explored 3D scenes to be updated with comparable accuracy and significantly reduced flight time. 
To summarize, the main contributions of our work are:
\begin{itemize}
    \item To the best of our knowledge, this paper presents the first path planning algorithm explicitly designed for detecting and updating change areas in urban scenes, significantly reducing flight time and computational overhead.
    \item We introduce a changeability heuristic to evaluate the potential changes in a 3D urban scene, which serves as the basis for planning two distinct flight trajectories for scene updating.
    \item Through a combination of statistical priors and real-time decision-making, our approach achieves high-quality updates with significantly reduced resource consumption, making it scalable and effective for large and complex urban environments.
\end{itemize}

\section{Related Works}
\label{sec:related_works}

Aerial path planning has been extensively studied in the context of urban scene reconstruction and exploration~\cite{fan2016automated,furukawa2010towards,hornung2008image,mauro2014unified,wu2014quality, kuang2020real}. As a complete system for large-scale urban scene change exploration, our approach involves three key technical aspects: image change detection, real-time scene exploration, and aerial path planning for online urban reconstruction. In this section, we review relevant works that lay the foundation for our approach.

\subsection{Aerial Image Change Detection}

Aerial image change detection is crucial for unmanned aerial vehicles (UAVs) to identify scene variations during flight, supporting subsequent analysis and reconstruction of altered areas. 
A closely related field is remote sensing change detection. 
Traditional hand-crafted methods\cite{bay2006surf,hornung2008image,rublee2011orb} compute local features and perform matching using predefined heuristic algorithms, which are unsuitable for images with low overlap or significant viewpoint variations. Learning-based methods~\cite{tian2017l2,mishchuk2017working,dusmanu2019d2,revaud2019r2d2,tyszkiewicz2020disk, fang2023changer, dong2024changeclip, varghese2018changenet, bertinetto2016fully}
have significantly improved the extraction of local image features through data-driven approaches. 
However, these methods typically rely on domain-specific datasets for training, resulting in poor generalization to in-the-wild data, particularly for aerial imagery, where complex scene variations pose significant challenges.
In our paper, we use GIM~\cite{shen2024gim} to achieve image matching with strong generalization and high performance on aerial images. GIM employs a self-training framework that leverages the rich data from internet videos, significantly enhancing the model's matching performance across diverse domains. This makes it more suitable for multi-view, multi-scale matching tasks in aerial imagery.


\subsection{Online and Real-Time Exploration}

Scene exploration has been a longstanding research topic. Ensuring all areas are observed during exploration is often referred to as the Complete Coverage Path Planning (CCPP) problem \cite{zelinsky1993planning}. A family of approaches inspired by the rapidly exploring random tree (RRT) \cite{bircher2016receding, selin2019efficient, schmid2020efficient} address this challenge by prioritizing online exploration of unknown regions. While effective in expanding known regions, these methods do not optimize for the quality of observations. Frontier-based method~\cite{cieslewski2017rapid} designs a boundary-driven system to enhance exploration efficiency for fast exploration. Unlike frontier-based methods, which are limited to high-speed flight applications, sampling-based approaches can be effectively integrated with a wide range of optimization objectives~\cite{karaman2011sampling}.
\citet{liu2021aerial} tackled this limitation by generating and updating image acquisition paths in real time through predefined trajectory patterns on coarse scene proxies. However, this approach, optimized for reconstructing entire scenes without prior knowledge, becomes redundant in scenarios focusing on localized change detection rather than comprehensive coverage. \citet{xu2017autonomous} proposed a time-varying tensor field method, which is well-suited for smooth path planning in ground robots that require obstacle avoidance. However, computing time-varying tensor fields is expensive hence its applicability to aerial drone path planning is limited in our context.

\subsection{Aerial Path Planning for Urban Reconstruction}

The reconstruction of urban environments using aerial imagery has attracted significant attention in recent years. 
To maximize information gain and minimize trajectory length, \cite{hepp2018plan3d,heng2015efficient,roberts2017submodular,koch2019automatic,smith2018aerial,zhou2020offsite} utilize geometric proxies of the scene to plan paths. 
\citet{huang2018active} explores unknown environments in real-time while simultaneously performing online 3D modeling. \citet{schmid2020efficient} and \citet{bircher2016receding} plan paths using sampling and optimization techniques to efficiently explore and reconstruct scenes in unknown environments. \citet{smith2018aerial} introduced a heuristically designed reconstructability measure to predict reconstruction quality. A continuous optimization method was subsequently developed to maximize the reconstructability of all points in the scene. Extending this idea, \citet{zhou2020offsite} proposed a MAXIMIN optimization approach to select viewpoints, ensuring maximum reconstructability under a constrained number of viewpoints. Similarly, \citet{zhang2021continuous} refined continuous path planning by emphasizing smooth trajectory designs (e.g. fewer sharp turns) that maintain effective scene coverage. These methods, however, primarily focus on offline optimization for high-quality reconstruction. To address the challenge of large-scale unknown environments in real-time, \citet{liu2021aerial} introduced a dual-purpose framework combining real-time exploration with offline reconstruction, achieving a balance between adaptability and planning. Building on this, \citet{liu2022learning} employed learning-based reconstructability metrics to optimize data acquisition based on scene complexity. 

Despite their advances, these methods focus on reconstructing entire scenes, often leading to prolonged processing times (e.g., a week for larger scenes). Our approach diverges from traditional urban reconstruction methodologies by leveraging aerial path planning specifically for change detection. By incorporating semantic labels from historical models and assigning probabilistic measures to potential changes, our method emphasizes areas with a higher likelihood of structural modifications. This targeted planning strategy optimizes information gain for change detection, significantly reducing unnecessary data acquisition for static areas.
\begin{figure*}[t]
    \centering
    \includegraphics[width=\linewidth]{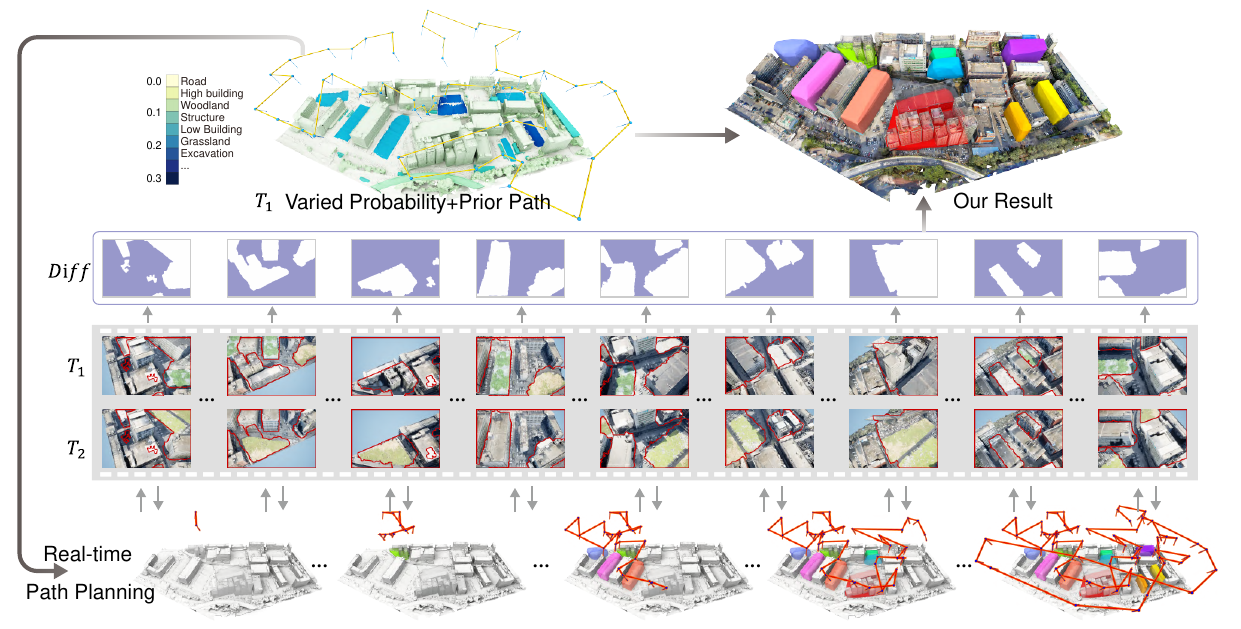}
    \caption{The input to our method is a labeled reconstructed model of $T_1$, where each semantic label is associated with a prior probability (Sec.~\ref{sec:changeability}). We first plan the \textit{prior path} by minimizing redundancy while maximizing potential \textit{changeability} and coverage (Sec.~\ref{sec:prior_path}). As the drone flies along the prior path and detects a change area, the \textit{real-time} path planning (Sec.~\ref{sec:realtime_path}) is triggered to explore the entire target change area by analyzing the \textit{diff} images. The output of our system is a set of convex hulls representing all detected change areas. }
    \label{fig:overview}
\end{figure*}

\section{Overview}
\label{sec:overview}

This work aims to optimize a UAV flight trajectory for detecting changes in a previously explored and reconstructed urban site. The initial exploration occurred at time $T_1$, and the new process takes place at time $T_2$. Prior knowledge from $T_1$ includes the reconstructed scene model with textures and labels~\cite{yang2023urbanbis}. Additionally, we utilize prior probability statistics from the WUSU dataset~\cite{shi2023openWUSU}, which show that buildings have a low probability of change, whereas open areas exhibit higher variability.

To identify and explore change areas on-the-fly, we address two key challenges: (1) designing an initial trajectory based on prior knowledge to maximize the exploration of potential change areas; and (2) efficiently exploring identified change areas. The output of our system is the convex hulls of all detected change regions. Addressing these challenges involves novel approaches, as our system is the first to tackle this specific problem.  

Our solution introduces two-step aerial paths for the change exploration process: the \textit{prior path} (see Sec.~\ref{sec:prior_path}) and the \textit{real-time path} (see Sec.~\ref{sec:realtime_path}). The prior path, generated using only the $T_1$ knowledge, serves as the initial trajectory for the UAV. 
The drone initially follows the prior path, but once a change area is detected, the real-time path planning process is triggered to generate a new path for the drone to detect the target change area. This new path is generated through an online next-best-view process for ensuring a thorough exploration of the target change area. 
Once exploration of the changed target is finished, the UAV resumes the remaining views generated during prior path planning to continue exploring the remaining areas. 
If additional change areas are detected along the path, they are queued for subsequent exploration, ensuring a comprehensive identification and analysis of all change areas.

An overview of our method is shown in Fig.~\ref{fig:overview}.
In the following section, we first introduce common components, such as candidate view generation and changeability heuristics. Then we will introduce the details of prior and real-time path planning in Sec.~\ref{sec:prior_path} and Sec.~\ref{sec:realtime_path}, respectively.

\subsection{Candidate View Generation}
\label{sec:candidate_view_gen}

\begin{wrapfigure}{r}{4cm}
\vspace{-3.5mm}
  \hspace*{-4mm}
  \centerline{
  \includegraphics[width=4cm]{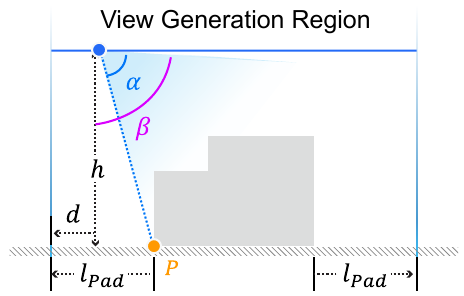}
  }
  \vspace*{-3.5mm}
\end{wrapfigure}
Candidate views for both the prior and the real-time paths are designed to sufficiently observe all the target surfaces $\target$ in a predefined safe height plane $h$. For the prior path, $\target = \targetprior$ corresponds to the reconstructed surface from $T_1$, while for the real-time path, $\target= \{\targetrealtime\}_{t=1}^M$ represents the convex hull of $M$ detected change areas, which are empty in the beginning and gradually expand if detected.
The region for generating new candidate views is determined by expanding the target, $\targetprior$ for the prior path and $\targetrealtime$ for the real-time path, with a padding margin $l_\text{pad}$. This padding ensures adequate coverage for all sampling points along the edges of the target's convex hull. Within this expanded region, Poisson disk sampling~\cite{corsini2012efficient} is applied to generate candidate views $\Views=\{\view_j\}$, where each candidate view $\view_j \in \mathbb{R}^6$ consists of its 3D position and orientation. 
To accommodate drones equipped with either multiple cameras (\eg five cameras) or single-camera systems, five candidate views are generated at the same position, each with a different orientation. This ensures compatibility and flexibility for varying drone configurations. Due to page limits, the detail formulation and parameters setup are provided in the Supplementary Materials.

\subsection{Scene Changeability Heuristics}
\label{sec:changeability}

Our candidate view generation process (Sec.~\ref{sec:candidate_view_gen}) produces a dense set of views to ensure sufficient coverage of the target surface $\target$. However, this introduces redundancy, making it necessary to prioritize and refine these views. For the prior path, the goal is to reduce redundancy while maintaining coverage. For the real-time path, the focus is on selecting the next-best view to explore the entire change area. To achieve this, we define a heuristic to quantify the importance of each view, guiding both planning tasks.

The heuristic aims to prioritize the detection of surface areas with high change probabilities. The target surface $\target$ is uniformly discretized into $N$ samples $\Samples = \{\sample_i, \probwrap(\sample_i) \}_{i=1}^N$, where each sample point $\sample_i$ is associated with a score $\probwrap(\sample_i)$, and
\begin{equation}
\probwrap(\sample_i) =
\begin{cases}
1 & \text{if } \sample_i \in \text{detected target change area } \targetrealtime, \\
\prob_i & \text{otherwise}.
\end{cases}
\label{eq:probwrap}
\end{equation}
If a sample is confirmed as part of a detected change area during real-time path planning, its score $\probwrap(\sample_i)=1$, indicating the highest changeability. Otherwise, $\probwrap(\sample_i) = \prob_i$ is the \textit{prior probability}. These probabilities are calculated using statistics from the WUSU dataset~\cite{shi2023openWUSU}, which includes 12 classes of urban structures. By analyzing the change rates across these classes, we map them to the 7 semantic labels in the UrbanBIS dataset~\cite{yang2023urbanbis}. Detailed statistics are provided in the Supplementary Material.

The \textit{visibility} function $\vis(\sample_i, \view_j)$ determines whether a sample $\sample_i$ is visible from a view $\view_j$:
\begin{equation}
\vis(\sample_i, \view_j) =
\begin{cases}
1 & \text{if } \sample_i \text{ is visible from } \view_j, \\
0 & \text{otherwise}.
\end{cases}
\end{equation}
The \textit{coverage} of a sample $\sample_i$ is the total number of views observing it:
\begin{equation}
\cov(\sample_i) = \sum_{\view_j \in \Views} \vis(\sample_i, \view_j).
\end{equation}
For real-time path planning, views $\Views$ are divided into a visited sequence $\Viewsvisited = \{\view_1, \dots, \view_j\}$ and unvisited $\Viewsunvisited$. The next-bestview is selected from $\Viewsunvisited$ to maximize the changeability. For prior path planning, all views are initially unvisited, $\Viewsvisited = \emptyset$.

The importance of a view $\view_j \in \Views$ is defined by the \textit{changeability} of the samples it observes. To achieve this, we first define the changeability of a sample $\sample_i$ with respect to a view $\view_j$ as:
\begin{equation}
f(\sample_i, \view_j) =
\begin{cases}
\omega \cdot q(\sample_i) \cdot \vis(\sample_i, \view_j) & \text{if } \view_j \in \Viewsunvisited, \\
-\gamma \cdot \vis(\sample_i, \view_j) & \text{if } \view_j \in \Viewsvisited,
\end{cases}
\label{eq:changeability_sv}
\end{equation}
where $\omega$ and $\gamma$ are scaling parameters that control the contribution of unvisited and visited views, respectively. The negative factor associated with visited views reflects diminishing returns: as a sample is observed by more visited views, its changeability stabilizes. This aligns with the idea that once a sample has been observed, its status as "changed" or "unchanged" becomes increasingly confirmed, yielding less new information with each additional observation. Consequently, the changeability decreases, highlighting the reduced benefit of redundant observations.

In subsequent sections, we define the specific importance of heuristics for both prior and real-time path planning, tailored to their respective goals. 

\section{Prior Path Planning}  
\label{sec:prior_path}  

Starting with the reconstructed scene $\targetprior$ from $T_1$, we derive the sample set $\Samplesprior = \{\sampleprior_i, \probwrap(\sampleprior_i)\}$, representing discretized target points with associated change probabilities. Using this input, we generate a dense set of candidate views $\Views$ over the target surface, as described in Sec.~\ref{sec:candidate_view_gen}.

The primary objective of prior path planning is to identify a minimal subset of views $\Viewsprior = \{\viewprior_j\} \subset \Views$ that reduces \textit{redundancy} while ensuring maximum \textit{coverage} of $\targetprior$. This objective draws inspiration from the MAXIMIN optimization framework~\cite{zhou2020offsite}. The trajectory connecting these selected views forms the prior aerial path.

To achieve this, we employ an iterative optimization process, similar to that in \citet{zhou2020offsite}, refining the view set by removing redundant views and maximizing target coverage. This optimization problem can be formulated as:
\begin{align}
\Viewsprior = \argmin_{\Viewsprior} \sum_{\viewprior_j \in \Viewsprior} r(\viewprior_j) +
\argmax_{\Viewsprior} \sum_{\sampleprior_i \in \Samplesprior} \cov(\sampleprior_i),
\end{align}
where the first term minimizes redundancy, and the second term maximizes coverage. Solving this optimization results in a refined view set $\Viewsprior$, which also serves as the initial candidate set for real-time path planning (see Sec.~\ref{sec:realtime_path}).
Throughout this process, all views are considered unvisited, i.e., $\Views = \Viewsunvisited$, and $\Viewsvisited = \emptyset$.

Unlike \citet{zhou2020offsite}, who define view redundancy based on scene \textit{reconstructability}~\cite{smith2018aerial}, we define redundancy in terms of \textit{changeability}. A view is considered \textit{redundant} if the samples it observes are either (1) already well-covered by other views or (2) have low changeability, making their observation less critical. Thus, the redundancy of a view can be quantified as the negative of its importance heuristic, $r(\viewprior_j) = -g(\viewprior_j)$, with higher redundancy corresponding to lower importance.

The importance heuristic of a view is derived from the changeability of the samples it observes. Changeability of a sample $\sampleprior_i \in \Samplesprior$ is determined based on the set of views observing it, $\Viewsobserve_i = \{\viewprior_j \mid \vis(\sampleprior_i, \viewprior_j) = 1\}$. It is defined as:
\begin{equation}
f(\sampleprior_i, \Viewsobserve_i) = \frac{1}{|\Viewsobserve_i|} \cdot f(\sampleprior_i, \viewprior_j),
\end{equation}
where $f(\sampleprior_i, \viewprior_j) = \beta \cdot \probprior_i$ for any $\viewprior_j \in \Viewsobserve_i$, as defined in Eq.~\eqref{eq:changeability_sv}. This formulation ensures that the changeability of a sample diminishes as it is observed by more views.
The importance heuristic of a view $\viewprior_j$ is then computed based on the samples it observes:
\begin{equation} 
g(\viewprior_j) = \sum_{\sampleprior_i \in \Samplesprior} f(\sampleprior_i, \Viewsobserve_i) \cdot \vis(\sampleprior_i, \viewprior_j) . 
\end{equation} 
To construct a continuous trajectory that visits all selected views in $\Viewsprior$, we frame the problem as a Traveling Salesman Problem (TSP), as in \citet{zhou2020offsite}. The solution ensures an efficient flight path that prioritizes areas with higher probabilities of change, based on prior statistics.
Further details of the prior path planning formulation are provided in the Supplementary Material due to space constraints.

\begin{figure*}[t]
    \centering
    \includegraphics[width=0.9\linewidth]{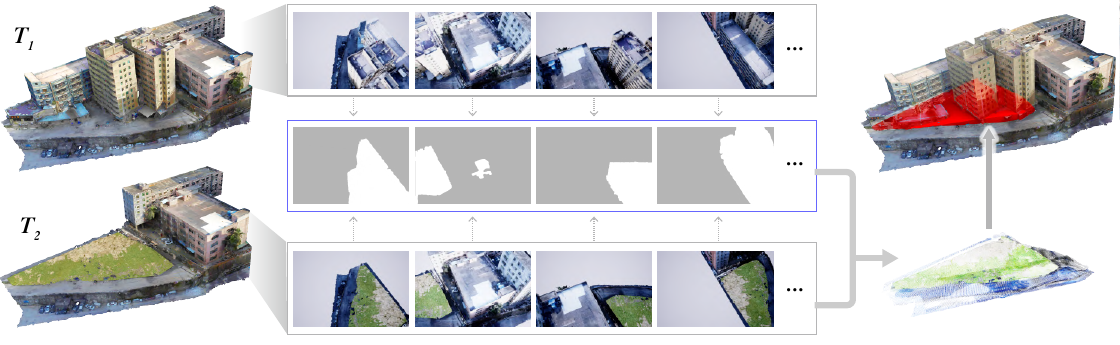}
    \caption{An example of our real-time path (detailed in Sec.~\ref{sec:realtime_path}) applied to a scene from UrbanBIS~\cite{yang2023urbanbis}. The $T_1$ reconstructed model (top left) serves as the prior, showing three existing buildings, while the $T_2$ model (bottom left) serves as the ground truth, where three buildings have been removed. As the drone detects changes in the scene, it dynamically adjusts its path by identifying differences in captured 2D images (middle). Subsequently, the detected point cloud (bottom right) and the convex hull of the change area (top right) are generated based on the updated real-time path.}
    \label{fig:realtime_path}
\end{figure*}

\section{Change Area Detection and Extraction}
\label{sec:realtime_change_detection}

At $T_2$, the drone sequentially follows the planned path $\{\view_{\text{start}}, \dots, \view_{\text{end}}\}$. At each designated view $\view_j$ in this sequence, it captures a 2D image $\image_j^{T_2}$ with the camera oriented in the direction $\viewdir_j$. Each captured image is then compared to a corresponding rendered image $\image_j^{T_1}$ generated from the $T_1$ reconstructed model with the same view and orientation.

To detect and identify areas of change at the image level, we use a learning-based image feature matching model, DKM~\cite{edstedt2023dkm}, enhanced by the generalized self-training framework GIM~\cite{shen2024gim}. This framework leverages large-scale internet videos for pretraining, ensuring robust generalization to diverse scenarios. Image regions that exhibit no feature matches between the image pair $(\image_j^{T_1}, \image_j^{T_2})$ are identified as change areas $\diffarea_j$.

To extract corresponding 3D information, we apply a pre-trained multi-view stereo (MVS) reconstruction method, 
DUSt3R~\cite{wang2024dust3r}.
The detected change areas $\diffarea_j$ are used as masks to exclude unchanged regions, yielding a 3D point cloud $\pointcloud^t$ that represents the identified change region. In cases where multiple change areas are detected, they are added to a processing queue and handled sequentially.

To ensure the performance using DUSt3R during path planning, the reconstruction of $\pointcloud^t$ incorporates only the nearest $8$ images from the recent view sequence $\{\view_{j-8}, \dots, \view_{j}\}$. However, this limitation may result in incomplete point clouds. To enhance the current target region $\pointcloud^t$ with newly generated point cloud data $\pointcloud^t_j$, their overlap is evaluated using the intersection-over-union (IoU) metric. If $\text{IoU}(\pointcloud^t, \pointcloud^t_j) < \phi$, where $\phi$ is a configurable similarity threshold, the updated target region is set as $\pointcloud^t = \pointcloud^t \cup \pointcloud^t_j$. Otherwise, the target region is replaced with the new point cloud, $\pointcloud^t = \pointcloud^t_j$.

At each view, the convex hull of the detected point cloud $\pointcloud^t$ is extracted as the detected change area. This convex hull serves as the target area  $\targetrealtime$ of interest for further exploration, informing real-time path planning decisions to determine the next-best view dynamically.

\section{Change Area Real-time Path Planning}
\label{sec:realtime_path}

Once a change area $\targetrealtime$ is identified, the system switches to real-time planning mode to dynamically and fully explore the region. If multiple change areas are detected, the system prioritizes the nearest change area relative to the drone's current position, selecting it as the next exploration target. Given the sequence of visited views $\Viewsvisited = \{\view_1, \dots, \view_j\}$ and the corresponding detected change target $\targetrealtime = \targetrealtime_j$ (if it exists), our goal for real-time planning is to select the next-best view $\view_{j+1}$ that maximizes exploration of $\targetrealtime$. This process dynamically updates the target area to $\targetrealtime = \targetrealtime_{j+1}$ as exploration progresses.
Similar to prior path planning in Sec.~\ref{sec:prior_path}, we first sample the target change area $\targetrealtime$ as $\Samplesrealtime = \{\samplerealtime_i, \probwrap(\samplerealtime_i)\}$, where $\probwrap(\samplerealtime_i) \equiv 1$ from Eq.~\ref{eq:probwrap}, and combine with all prior samples $\Samples = \Samplesrealtime \bigcup \Samplesprior$.
Let $\view_{j+1}$ represent a new unvisited view to be appended to the aerial path. To evaluate the contribution of a candidate view, we first consider the changeability of all samples that this view can observe. 
The changeability for a sample $\sample_i \in \Samples$ relative to the candidate next view $\view_{j+1} \in \Viewsunvisited$, considering the sequence of already visited views $\Viewsvisited$, is defined as: 
\begin{equation}
f(\sample_i, \view_{j+1} | \Viewsvisited) = f(\sample_i, \view_{j+1}) + \sum_{\view_k \in \Viewsvisited} f(\sample_i, \view_k).
\end{equation}
The total changeability gain, or importance heuristic, for a candidate view $\view_{j+1}$ is then calculated as the cumulative gain it provides, taking into account the information already acquired by the previously visited views:  
\begin{equation}
g(\view_{j+1} | \Viewsvisited) = \sum_{\sample_i \in \Samples} f(\sample_i,\view_{j+1} | \Viewsvisited).
\end{equation}

Candidate views for $\view_{j+1}$ are derived from two sources: (1) Unvisited views from the prior path $\Viewsprior\setminus\Viewsvisited$;
(2) A set of real-time generated views, $\Viewsrealtime$, created using Poisson disk sampling (see Sec.~\ref{sec:candidate_view_gen}). We filter the combined set to include only views that can observe at least one sample in $\Samplesrealtime$:
\begin{equation}
\Viewsunvisited = \{\view_{j+1} \mid \view_{j+1} \in (\Viewsprior\setminus\Viewsvisited) \cup \Viewsrealtime, \vis(\samplerealtime_i, \view_{j+1}) = 1, \, \exists \samplerealtime_i \in \Samplesrealtime\}.
\end{equation}
For each $\view_{j+1} \in \Viewsunvisited$, the changeability gain is computed in parallel. To prevent extreme trajectory deviations, the system selects the top $K$ candidates with the highest changeability gains and chooses the one closest to the current view $\view_j$ as the next-best view. In our experiments, $K$ is set to 10. The real-time path is then naturally constructed by sequentially connecting the visited path to the selected next-best view.  

The exploration of $\targetrealtime$ is considered complete once all samples $\samplerealtime_i \in \Samplesrealtime$ have been observed by the visited views $\Viewsvisited$. If multiple change areas are identified, the system proceeds to the next closest target $\target^{t+1}$ and repeats this process.  

After all detected change areas have been explored, the system revisits any remaining unobserved regions. Unvisited views from the prior path, $\Viewsprior \setminus \Viewsvisited$, are filtered to include only those capable of observing unvisited samples. These views are then connected into a continuous trajectory using a Traveling Salesman Problem (TSP) formulation, as in the prior path planning stage, to ensure efficient coverage of the remaining scene.  
By dynamically exploring identified change areas and revisiting unexplored regions, the real-time path planning strategy ensures comprehensive and efficient scene updates.

\begin{figure*}[ht]
    \centering
    \includegraphics[width=0.9\linewidth]{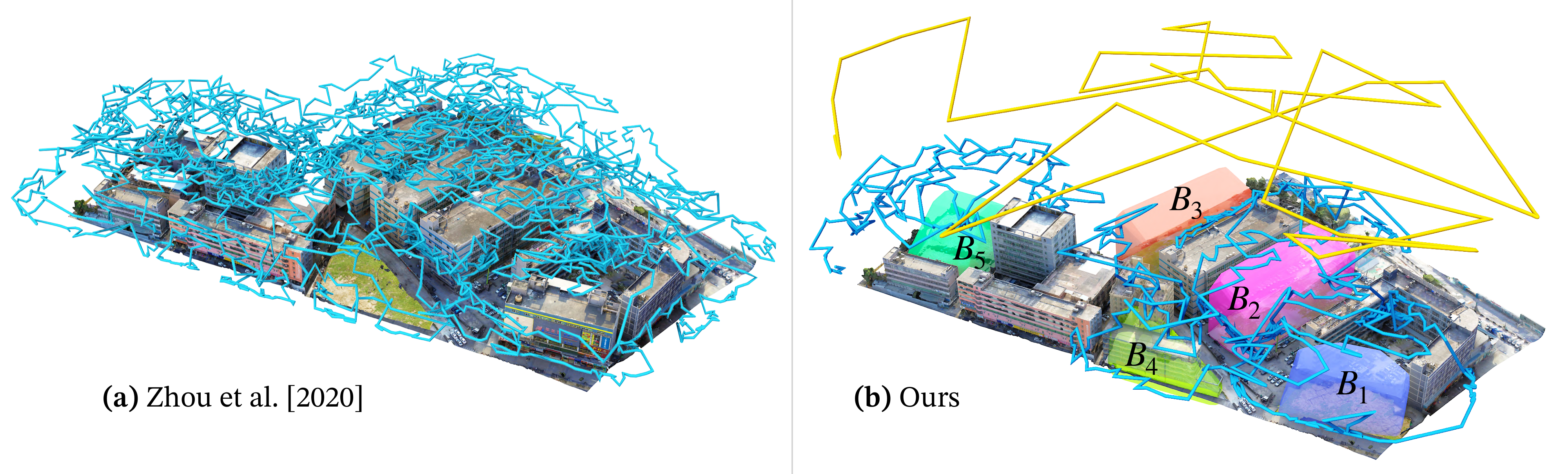}
    \caption{Visual comparison of the efficiency of our method (b) versus \citet{zhou2020offsite} (a). The experimental setup is detailed in Sec.~\ref{sec:exp_efficiency}, and the corresponding statistics are presented in Table~\ref{tab:exp_efficiency}. Our method demonstrates significantly improved efficiency, making it well-suited for downstream tasks such as reconstruction. }
    \label{fig:fig_efficiency}
\end{figure*}

\section{Results and Evaluation}  
\label{sec:exp}  

To validate our method, we performed a series of experiments using real-world environments from the UrbanBIS~\cite{yang2023urbanbis} dataset. Our path planning is implemented in C++ and run on a computer with a Intel Core i9-13900K Processor and a RTX 4090 GPU. For drone flight simulation and image capture, we utilized \textit{Unreal Engine}\footnote{https://www.unrealengine.com/} and \textit{Colosseum}\footnote{https://github.com/CodexLabsLLC/Colosseum}, which provided realistic and controllable simulation environments. The captured images were used to reconstruct 3D meshes of the scenes with the commercial software \textit{RealityCapture}\footnote{https://www.capturingreality.com}. To ensure fairness and consistency across all experiments, we used the default settings of RealityCapture, maintaining uniformity in the reconstruction process. The safe-height plane for drone flight was set to $h=120$ meters in all experiments. We use $\omega = 3$ and $\gamma = 2$ for Eq.~\eqref{eq:changeability_sv}, and set $\phi=0.3$ in Sec.~\ref{sec:realtime_change_detection}.

While scene updates have numerous practical applications, obtaining real-world datasets for this purpose remains challenging. Consequently, most of our test data involves manually updating a $T_1$ scene to generate a corresponding $T_2$ scene for evaluation. However, we also include an experiment comparing real-world $T_1$ and $T_2$ scenes to demonstrate the effectiveness of our method in practical scenarios in Sec.~\ref{sec:exp_large_real}. 

\begin{figure}[h]
    \centering
    \includegraphics[width=\linewidth]{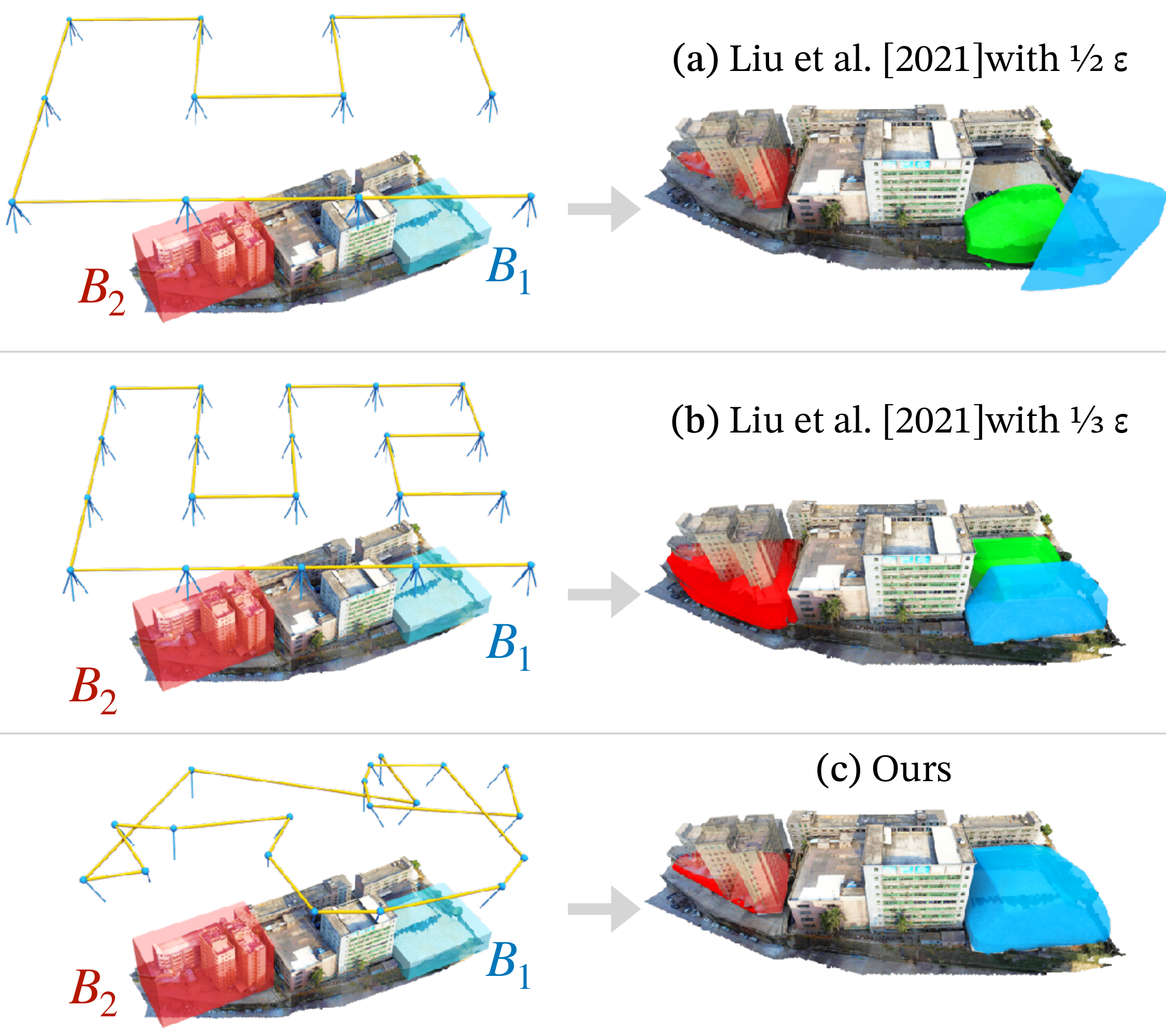}
    \caption{Qualitative comparison of path planning methods for detecting and extracting change areas. Refer to Sec.~\ref{sec:exp_change_exp} for experimental details. The \textit{Region-Division} method~\cite{liu2021aerial} is evaluated with two grid sizes: $\frac{1}{2} \epsilon$ (a) and $\frac{1}{3} \epsilon$ (b). 
    Our approach demonstrates accurate detection of change areas while requiring fewer viewpoints and shorter trajectories, with quantitative metrics provided in Table~\ref{tab:exp_change_exp}. }
    \label{fig:fig_exp_explore}
\end{figure}

\subsection{Change Area Exploration}
\label{sec:exp_change_exp}

Since no existing path planning methods are specifically designed for change detection, we compare our approach with the scene exploration method \textit{Region-Division}~(RD)~\cite{liu2021aerial}. This baseline allows the drone to fly over a region of size $\epsilon$, which is pre-divided into grids. The flight path is predefined to ensure all grids are explored. For each grid, the drone captures $4$ images from different directions, and these images are compared with those captured at $T_1$. To detect and extract change areas, we employ the same methodology described in Sec.~\ref{sec:realtime_change_detection}, constructing the convex hull of the detected change regions.
To ensure a fair comparison, we evaluate the Region-Division method with two grid sizes, $\frac{1}{2} \epsilon$ and $\frac{1}{3} \epsilon$. We assess the methods on a real-world scene from UrbanScene3D~\cite{2022urbanscene3d} dataset and evaluate the quality of the planned change detection paths using the following metrics:
(1) Intersection over Union (IoU) to measures the quality of the extracted change area compared to the ground truth change area at $T_2$; (2) Number of Viewpoints (\#Views) indicates the total number of views visited by the drone to detect changes; (3) Trajectory Length (Path Len) in meters to quantify the total distance traveled by the drone during the exploration.
Since RD~\cite{liu2021aerial} follows a pre-defined grid path, their planning time is not included in the comparison. For this scene, our method demonstrates an \textit{average real-time path planning time} of $0.085$ seconds per step. It is the average computation time per frame for selecting the next-best view, excluding the time consumed by replaceable components such as DUSt3R for reconstruction. A summary table of this time metric for all test scenes is given in the Supplementary Material.
The statistics of change area exploration are summarized in Table~\ref{tab:exp_change_exp}, and the visual comparison is shown in Fig.~\ref{fig:fig_exp_explore}. The evaluated scene features two buildings with significant changes: building $B_1$, which was constructed after $T_1$, and building $B_2$, which existed at $T_1$ but was removed by $T_2$.
Our results reveal that while larger grid sizes result in shorter flight paths, they compromise the detection and extraction performance of change areas. In contrast, our method achieves higher IoU and superior change area detection quality, requiring fewer viewpoints and images. These findings underscore the efficiency and accuracy of our approach in change detection tasks.

\begin{table}[h]
\small
\caption{Quantitative comparison of path planning strategies for exploring and extracting change areas. Details of the setup are provided in Sec.~\ref{sec:exp_change_exp} and the visual comparison is given in Fig.~\ref{fig:fig_exp_explore}.
Metrics include the Intersection over Union (IoU) for each changed building, the number of viewpoints (\#Views), and the trajectory length (Path Len) in meters. The best results are highlighted in bold. }
\begin{center}
\scalebox{1}{
\begin{tabular}{c||c|c||c||c}
\hline
Method & IoU $B_1$ $\uparrow$ & IoU $B_2$ $\uparrow$ & $\#$Views $\downarrow$ & Path Len [m] $\downarrow$ \\
\hline
RD-$\frac{1}{2} \epsilon$ & 0.201 & 0.029 & 48 & \textbf{973} \\    
\hline
RD-$\frac{1}{3} \epsilon$ & 0.801 & 0.059 & 80 & 1,121 \\    
\hline
Ours & \textbf{0.840} & \textbf{0.253} & \textbf{19} & 994 \\    
\hline
\end{tabular}}
\end{center}
\label{tab:exp_change_exp}
\end{table}

\begin{table}[t!]
\small
\caption{Quantitative comparison of two reconstructed scenes using either the \textit{Coarse Proxy} approach~\cite{zhou2020offsite} or the convex hulls generated by our method as input to the same downstream reconstruction approach proposed by~\citet{zhou2020offsite}. Visual results are presented in Fig.~\ref{fig:exp_recon}, and the definitions of the two evaluation metrics are provided in Sec.~\ref{sec:exp_change_recon}. The convex hulls obtained from our method can effectively replace coarse proxies in terms of reconstruction quality.}
\begin{center}
\scalebox{0.9}{
\begin{tabular}{c||c||c|c||c|c}
\hline
Scene & Input Method & 
\begin{tabular}{@{}c@{}}Error[m] \\ $90\%$ $\downarrow$ \end{tabular} & 
\begin{tabular}{@{}c@{}}Error[m] \\ $95\%$ $\downarrow$ \end{tabular} & 
\begin{tabular}{@{}c@{}}Comp[\%] \\ $0.05$m $\uparrow$ \end{tabular} & 
\begin{tabular}{@{}c@{}}Comp[\%] \\ $0.075$m $\uparrow$ \end{tabular} \\
\hline
\multirow{2}{*}{\hfil Scene 1} 
& Coarse Proxy & 	$0.122$ &	$0.197$ & $\mathbf{63.09}$ & $\mathbf{76.81}$ \\
& Ours & $\mathbf{0.110}$ & $\mathbf{0.171}$ & $60.86$ &	$75.77$ \\
\hline

\multirow{2}{*}{\hfil Scene 2} 
& Coarse Proxy & $\mathbf{0.103}$ & $0.187$ & $63.47$ & $79.21$ \\
& Ours & $0.107$ &	$\mathbf{0.172}$ &	$\mathbf{67.76}$ & $\mathbf{80.58}$ \\
\hline
\end{tabular}}
\end{center}
\label{tab:exp_recon}
\end{table}

\begin{table}[h!]
\small
\caption{Statistics comparing path efficiency between the method from \citet{zhou2020offsite} and our proposed scene update approach. Our scene update path (Ours-All) is divided into two components: the change detection path (Ours-CD) and the reconstruction path (Ours-Recon). For consistency, the reconstruction path is generated using the same method as \citet{zhou2020offsite}. 
The corresponding visual comparison is given in Fig.~\ref{fig:fig_efficiency} and detail described in Sec.~\ref{sec:exp_efficiency}.}
\begin{center}
\small
\scalebox{1}{
\begin{tabular}{c||c||c}
\hline
Method & $\#$Views $\downarrow$ & Path Len [m] $\downarrow$ \\
\hline
\citet{zhou2020offsite} & 4,303 & 22,031 \\
\hline
Ours-CD & 36 & 2,348 \\
Ours-Rcon-$B_1$ & 224 & 1,260 \\
Ours-Rcon-$B_2$ & 320 & 1,648 \\
Ours-Rcon-$B_3$ & 311 & 1,800 \\
Ours-Rcon-$B_4$ & 94 & 936 \\
Ours-Rcon-$B_5$ & 259 & 1,680 \\
Ours-All & \textbf{1,244} & \textbf{9,672} \\
\hline
\end{tabular}}
\end{center}
\label{tab:exp_efficiency}
\end{table}


\begin{figure*}[ht]
    \centering
    \includegraphics[width=\linewidth]{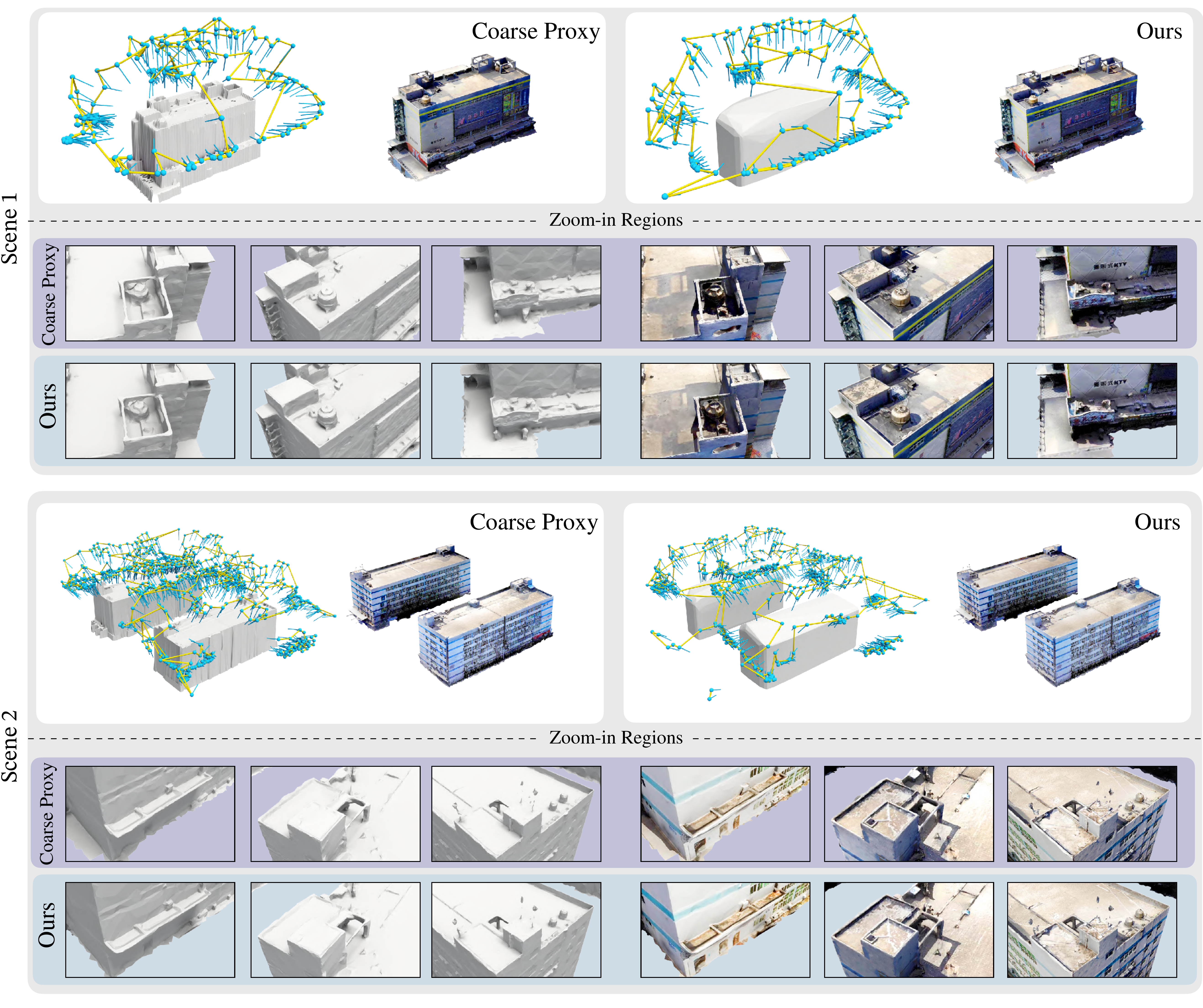}
    \caption{Visual comparison of reconstructed buildings using either the \textit{coarse proxy} approach from~\cite{zhou2020offsite} or the convex hulls generated by our method. Quantitative statistics corresponding to these results are provided in Table~\ref{tab:exp_recon}. 
    Reconstructions are shown for two different scenes, we provide more visual and quantitative comparisons in the Supplementary Material.
    }
    \label{fig:exp_recon}
\end{figure*}

\subsection{Scene Update Efficiency}
\label{sec:exp_efficiency}

To evaluate the efficiency of our method, we compare the planned trajectories for scene reconstruction. Once the drone completes image capture along both the prior and real-time paths, the convex hulls of all detected change areas are generated. These convex hulls can be directly utilized in downstream tasks such as 3D reconstruction, where the detected change areas serve as focused targets. The reconstruction process can seamlessly integrate with existing planners~\cite{zhou2020offsite, zhang2021continuous, smith2018aerial, liu2021aerial}. 
For our experiments, we utilize the reconstruction planner from \citet{zhou2020offsite} to plan the reconstruction path specifically for the detected change areas. We compare our approach against the conventional method of using \citet{zhou2020offsite} to re-explore and reconstruct the entire scene. Visual comparisons are provided in Fig.~\ref{fig:fig_efficiency}, while quantitative metrics are summarized in Table~\ref{tab:exp_efficiency}. 
The average real-time path (Ours-CD in Table~\ref{tab:exp_efficiency}) planning time is $0.063$ seconds per step for this scene.

The test scene contains $5$ distinct change areas ($B_1$ to $B_5$), all of which are correctly detected by our method. Compared to the traditional approach, our method achieves significant efficiency gains, reducing trajectory length by $52\%$ and the number of viewpoints by $71\%$. These results demonstrate the advantage of our targeted approach in detecting and updating change areas, enabling efficient and scalable updates for large-scale urban scenes.

\begin{figure*}[ht]
    \centering
    \includegraphics[width=\linewidth]{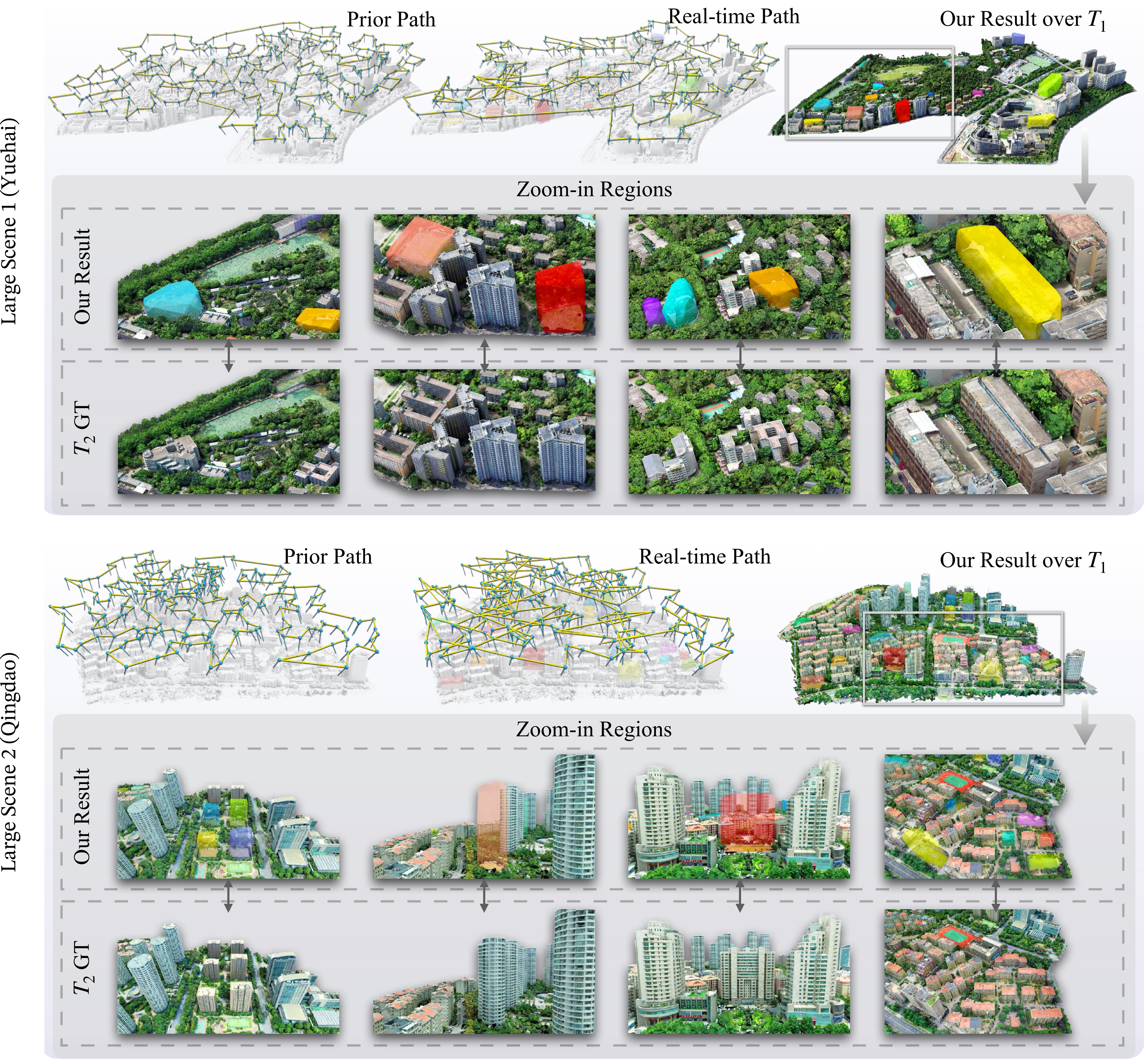}
    \caption{Qualitative results for two large-scale scenes from the UrbanBIS~\cite{yang2023urbanbis} dataset, as discussed in Sec.~\ref{sec:exp_large_real}. Detailed statistics are provided in the Supplementary Material due to space constraints}
    \label{fig:exp_large}
\end{figure*}

\begin{figure*}[ht]
    \centering
    \includegraphics[width=\linewidth]{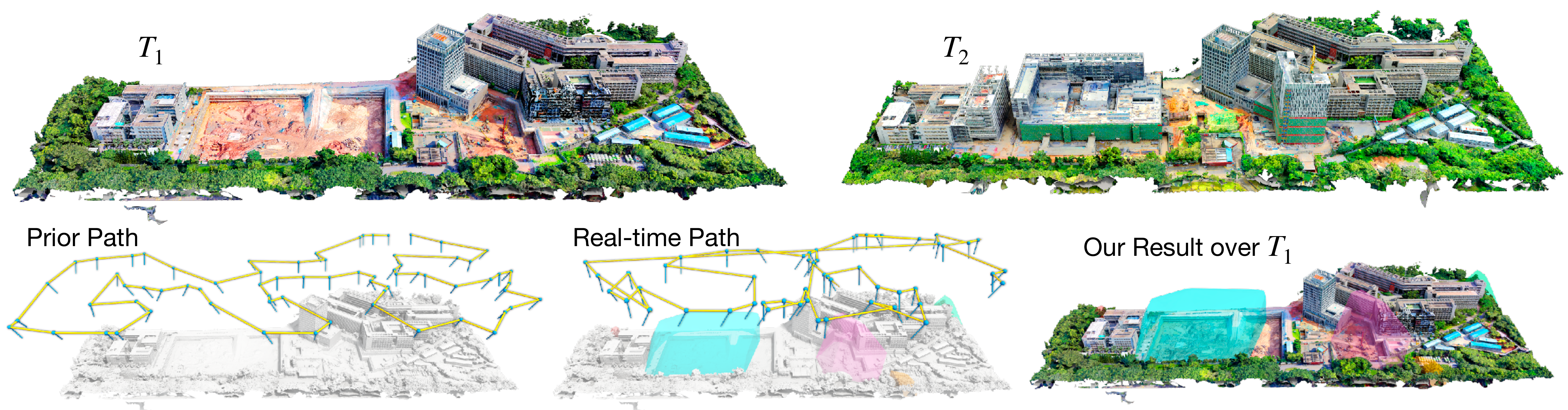}
    \caption{We test our method on one real-world scene where $T_1$ was scanned in 2021 and $T_2$ was scanned in 2023. The detail analysis of this scene is shown in the Supplementary Material. }
    \label{fig:exp_real}
\end{figure*}

\subsection{Change Area Reconstruction}
\label{sec:exp_change_recon}

We present both quantitative and qualitative comparisons of reconstruction quality using our convex hulls as input versus the \textit{coarse proxies} from~\cite{zhou2020offsite}. The results are summarized in Table~\ref{tab:exp_recon} and illustrated in Fig.~\ref{fig:exp_recon}. The coarse proxy from~\cite{zhou2020offsite} is a 2.5D proxy generated using a map and satellite image. 

Following the methodology of \citet{smith2018aerial}, we evaluate the quality of each reconstructed change area using two key metrics: \textit{Error} and \textit{Completeness}. The \textit{Error} metric measures how closely the reconstructed mesh approximates the ground truth, with smaller values indicating higher reconstruction quality. Conversely, the \textit{Completeness} metric assesses how well the reconstructed mesh covers the ground truth, with higher percentages indicating better coverage. Details of the computation for these metrics are provided in the Supplementary Materials due to space constraints. Our method achieves reconstruction quality comparable to existing approaches.

\subsection{Tests on Large and Real-World Scenes}  
\label{sec:exp_large_real}  

We evaluate our method on two extensive urban scenes from the UrbanBIS~\cite{yang2023urbanbis} dataset, as illustrated in Fig.~\ref{fig:exp_large}. For these tests, certain areas within the scenes were manually updated to simulate changes. Detailed statistics for these two scenes are included in the Supplementary Material.  
Additionally, we assess our method on a real-world scene where $T_1$ was scanned in 2021 and $T_2$ in 2023, as shown in Fig.~\ref{fig:exp_real}. A comprehensive analysis of this real-world scenario is provided in the Supplementary Material.

\section{Limitations and future work}
\label{sec:limitations}

In this paper, we proposed a method for detecting and exploring change areas in urban scenes using prior and real-time aerial path planning. 
Experimental results demonstrate the feasibility and effectiveness of our system. However, there are several limitations that highlight avenues for future work. 
First, our current implementation uses a fixed flight altitude of $h=120$ meters. This leads to challenges in scenarios with buildings of varying heights: taller buildings may be viewed too closely, resulting in fragmented convex hulls, while shorter buildings may not be captured clearly if the altitude is increased. To address this, we plan to develop an adaptive altitude mechanism that adjusts the flight height dynamically based on building heights and scene structure.
Second, while we use the DUSt3R~\cite{wang2024dust3r} method for multi-view point cloud reconstruction, its processing speed is relatively slow, which limits the real-time convex hull extraction speed. 
In the future, we aim to replace DUSt3R with a faster and more robust reconstruction algorithm as advancements in real-time methods emerge.
Third, our algorithm is primarily designed for urban building scenes and is not suitable for forest scenes or large-scale changes like those caused by earthquakes.


\begin{acks}
We thank all the anonymous reviewers for their insightful comments. Thanks also go to Yilin Liu and Yuan Liu for helpful discussions. This work was supported in parts by Guangdong S\&T Program (2024B01015004), NSFC (U21B2023), DEGP Innovation Team (2022KCXTD025), Shenzhen Science and Technology Program (KJZD20240903100022028, KQTD20210811090044003, RCJC20200714 114435012), and Scientific Development Funds from SZU.
\end{acks}

\bibliographystyle{ACM-Reference-Format}
\bibliography{reference}


\begin{thebibliography}{43}


\ifx \showCODEN    \undefined \def \showCODEN     #1{\unskip}     \fi
\ifx \showDOI      \undefined \def \showDOI       #1{#1}\fi
\ifx \showISBNx    \undefined \def \showISBNx     #1{\unskip}     \fi
\ifx \showISBNxiii \undefined \def \showISBNxiii  #1{\unskip}     \fi
\ifx \showISSN     \undefined \def \showISSN      #1{\unskip}     \fi
\ifx \showLCCN     \undefined \def \showLCCN      #1{\unskip}     \fi
\ifx \shownote     \undefined \def \shownote      #1{#1}          \fi
\ifx \showarticletitle \undefined \def \showarticletitle #1{#1}   \fi
\ifx \showURL      \undefined \def \showURL       {\relax}        \fi
\providecommand\bibfield[2]{#2}
\providecommand\bibinfo[2]{#2}
\providecommand\natexlab[1]{#1}
\providecommand\showeprint[2][]{arXiv:#2}

\bibitem[Bay(2006)]%
        {bay2006surf}
\bibfield{author}{\bibinfo{person}{Herbert Bay}.} \bibinfo{year}{2006}\natexlab{}.
\newblock \showarticletitle{Surf: Speeded up robust features}.
\newblock \bibinfo{journal}{\emph{Computer Vision—ECCV}} (\bibinfo{year}{2006}).
\newblock


\bibitem[Bertinetto et~al\mbox{.}(2016)]%
        {bertinetto2016fully}
\bibfield{author}{\bibinfo{person}{Luca Bertinetto}, \bibinfo{person}{Jack Valmadre}, \bibinfo{person}{Joao~F Henriques}, \bibinfo{person}{Andrea Vedaldi}, {and} \bibinfo{person}{Philip~HS Torr}.} \bibinfo{year}{2016}\natexlab{}.
\newblock \showarticletitle{Fully-convolutional siamese networks for object tracking}. In \bibinfo{booktitle}{\emph{Computer vision--ECCV 2016 workshops: Amsterdam, the Netherlands, October 8-10 and 15-16, 2016, proceedings, part II 14}}. Springer, \bibinfo{pages}{850--865}.
\newblock


\bibitem[Bircher et~al\mbox{.}(2016)]%
        {bircher2016receding}
\bibfield{author}{\bibinfo{person}{Andreas Bircher}, \bibinfo{person}{Mina Kamel}, \bibinfo{person}{Kostas Alexis}, \bibinfo{person}{Helen Oleynikova}, {and} \bibinfo{person}{Roland Siegwart}.} \bibinfo{year}{2016}\natexlab{}.
\newblock \showarticletitle{Receding horizon" next-best-view" planner for 3d exploration}. In \bibinfo{booktitle}{\emph{2016 IEEE international conference on robotics and automation (ICRA)}}. IEEE, \bibinfo{pages}{1462--1468}.
\newblock


\bibitem[Cieslewski et~al\mbox{.}(2017)]%
        {cieslewski2017rapid}
\bibfield{author}{\bibinfo{person}{Titus Cieslewski}, \bibinfo{person}{Elia Kaufmann}, {and} \bibinfo{person}{Davide Scaramuzza}.} \bibinfo{year}{2017}\natexlab{}.
\newblock \showarticletitle{Rapid exploration with multi-rotors: A frontier selection method for high speed flight}. In \bibinfo{booktitle}{\emph{2017 IEEE/RSJ International Conference on Intelligent Robots and Systems (IROS)}}. IEEE, \bibinfo{pages}{2135--2142}.
\newblock


\bibitem[Corsini et~al\mbox{.}(2012)]%
        {corsini2012efficient}
\bibfield{author}{\bibinfo{person}{Massimiliano Corsini}, \bibinfo{person}{Paolo Cignoni}, {and} \bibinfo{person}{Roberto Scopigno}.} \bibinfo{year}{2012}\natexlab{}.
\newblock \showarticletitle{Efficient and flexible sampling with blue noise properties of triangular meshes}.
\newblock \bibinfo{journal}{\emph{IEEE transactions on visualization and computer graphics}} \bibinfo{volume}{18}, \bibinfo{number}{6} (\bibinfo{year}{2012}), \bibinfo{pages}{914--924}.
\newblock


\bibitem[Dong et~al\mbox{.}(2024)]%
        {dong2024changeclip}
\bibfield{author}{\bibinfo{person}{Sijun Dong}, \bibinfo{person}{Libo Wang}, \bibinfo{person}{Bo Du}, {and} \bibinfo{person}{Xiaoliang Meng}.} \bibinfo{year}{2024}\natexlab{}.
\newblock \showarticletitle{ChangeCLIP: Remote sensing change detection with multimodal vision-language representation learning}.
\newblock \bibinfo{journal}{\emph{ISPRS Journal of Photogrammetry and Remote Sensing}}  \bibinfo{volume}{208} (\bibinfo{year}{2024}), \bibinfo{pages}{53--69}.
\newblock


\bibitem[Dusmanu et~al\mbox{.}(2019)]%
        {dusmanu2019d2}
\bibfield{author}{\bibinfo{person}{Mihai Dusmanu}, \bibinfo{person}{Ignacio Rocco}, \bibinfo{person}{Tomas Pajdla}, \bibinfo{person}{Marc Pollefeys}, \bibinfo{person}{Josef Sivic}, \bibinfo{person}{Akihiko Torii}, {and} \bibinfo{person}{Torsten Sattler}.} \bibinfo{year}{2019}\natexlab{}.
\newblock \showarticletitle{D2-net: A trainable cnn for joint description and detection of local features}. In \bibinfo{booktitle}{\emph{Proceedings of the ieee/cvf conference on computer vision and pattern recognition}}. \bibinfo{pages}{8092--8101}.
\newblock


\bibitem[Edstedt et~al\mbox{.}(2023)]%
        {edstedt2023dkm}
\bibfield{author}{\bibinfo{person}{Johan Edstedt}, \bibinfo{person}{Ioannis Athanasiadis}, \bibinfo{person}{M{\aa}rten Wadenb{\"a}ck}, {and} \bibinfo{person}{Michael Felsberg}.} \bibinfo{year}{2023}\natexlab{}.
\newblock \showarticletitle{DKM: Dense kernelized feature matching for geometry estimation}. In \bibinfo{booktitle}{\emph{Proceedings of the IEEE/CVF Conference on Computer Vision and Pattern Recognition}}. \bibinfo{pages}{17765--17775}.
\newblock


\bibitem[Fan et~al\mbox{.}(2016)]%
        {fan2016automated}
\bibfield{author}{\bibinfo{person}{Xinyi Fan}, \bibinfo{person}{Linguang Zhang}, \bibinfo{person}{Benedict Brown}, {and} \bibinfo{person}{Szymon Rusinkiewicz}.} \bibinfo{year}{2016}\natexlab{}.
\newblock \showarticletitle{Automated view and path planning for scalable multi-object 3D scanning}.
\newblock \bibinfo{journal}{\emph{ACM Transactions on Graphics (TOG)}} \bibinfo{volume}{35}, \bibinfo{number}{6} (\bibinfo{year}{2016}), \bibinfo{pages}{1--13}.
\newblock


\bibitem[Fang et~al\mbox{.}(2023)]%
        {fang2023changer}
\bibfield{author}{\bibinfo{person}{Sheng Fang}, \bibinfo{person}{Kaiyu Li}, {and} \bibinfo{person}{Zhe Li}.} \bibinfo{year}{2023}\natexlab{}.
\newblock \showarticletitle{Changer: Feature interaction is what you need for change detection}.
\newblock \bibinfo{journal}{\emph{IEEE Transactions on Geoscience and Remote Sensing}}  \bibinfo{volume}{61} (\bibinfo{year}{2023}), \bibinfo{pages}{1--11}.
\newblock


\bibitem[Furukawa et~al\mbox{.}(2010)]%
        {furukawa2010towards}
\bibfield{author}{\bibinfo{person}{Yasutaka Furukawa}, \bibinfo{person}{Brian Curless}, \bibinfo{person}{Steven~M Seitz}, {and} \bibinfo{person}{Richard Szeliski}.} \bibinfo{year}{2010}\natexlab{}.
\newblock \showarticletitle{Towards internet-scale multi-view stereo}. In \bibinfo{booktitle}{\emph{2010 IEEE computer society conference on computer vision and pattern recognition}}. IEEE, \bibinfo{pages}{1434--1441}.
\newblock


\bibitem[Heng et~al\mbox{.}(2015)]%
        {heng2015efficient}
\bibfield{author}{\bibinfo{person}{Lionel Heng}, \bibinfo{person}{Alkis Gotovos}, \bibinfo{person}{Andreas Krause}, {and} \bibinfo{person}{Marc Pollefeys}.} \bibinfo{year}{2015}\natexlab{}.
\newblock \showarticletitle{Efficient visual exploration and coverage with a micro aerial vehicle in unknown environments}. In \bibinfo{booktitle}{\emph{2015 IEEE International Conference on Robotics and Automation (ICRA)}}. IEEE, \bibinfo{pages}{1071--1078}.
\newblock


\bibitem[Hepp et~al\mbox{.}(2018a)]%
        {hepp2018learn}
\bibfield{author}{\bibinfo{person}{Benjamin Hepp}, \bibinfo{person}{Debadeepta Dey}, \bibinfo{person}{Sudipta~N Sinha}, \bibinfo{person}{Ashish Kapoor}, \bibinfo{person}{Neel Joshi}, {and} \bibinfo{person}{Otmar Hilliges}.} \bibinfo{year}{2018}\natexlab{a}.
\newblock \showarticletitle{Learn-to-score: Efficient 3d scene exploration by predicting view utility}. In \bibinfo{booktitle}{\emph{Proceedings of the European conference on computer vision (ECCV)}}. \bibinfo{pages}{437--452}.
\newblock


\bibitem[Hepp et~al\mbox{.}(2018b)]%
        {hepp2018plan3d}
\bibfield{author}{\bibinfo{person}{Benjamin Hepp}, \bibinfo{person}{Matthias Nie{\ss}ner}, {and} \bibinfo{person}{Otmar Hilliges}.} \bibinfo{year}{2018}\natexlab{b}.
\newblock \showarticletitle{Plan3d: Viewpoint and trajectory optimization for aerial multi-view stereo reconstruction}.
\newblock \bibinfo{journal}{\emph{ACM Transactions on Graphics (TOG)}} \bibinfo{volume}{38}, \bibinfo{number}{1} (\bibinfo{year}{2018}), \bibinfo{pages}{1--17}.
\newblock


\bibitem[Hornung et~al\mbox{.}(2008)]%
        {hornung2008image}
\bibfield{author}{\bibinfo{person}{Alexander Hornung}, \bibinfo{person}{Boyi Zeng}, {and} \bibinfo{person}{Leif Kobbelt}.} \bibinfo{year}{2008}\natexlab{}.
\newblock \showarticletitle{Image selection for improved multi-view stereo}. In \bibinfo{booktitle}{\emph{2008 IEEE Conference on Computer Vision and Pattern Recognition}}. IEEE, \bibinfo{pages}{1--8}.
\newblock


\bibitem[Huang et~al\mbox{.}(2018)]%
        {huang2018active}
\bibfield{author}{\bibinfo{person}{Rui Huang}, \bibinfo{person}{Danping Zou}, \bibinfo{person}{Richard Vaughan}, {and} \bibinfo{person}{Ping Tan}.} \bibinfo{year}{2018}\natexlab{}.
\newblock \showarticletitle{Active image-based modeling with a toy drone}. In \bibinfo{booktitle}{\emph{2018 IEEE International Conference on Robotics and Automation (ICRA)}}. IEEE, \bibinfo{pages}{6124--6131}.
\newblock


\bibitem[Karaman and Frazzoli(2011)]%
        {karaman2011sampling}
\bibfield{author}{\bibinfo{person}{Sertac Karaman} {and} \bibinfo{person}{Emilio Frazzoli}.} \bibinfo{year}{2011}\natexlab{}.
\newblock \showarticletitle{Sampling-based algorithms for optimal motion planning}.
\newblock \bibinfo{journal}{\emph{The international journal of robotics research}} \bibinfo{volume}{30}, \bibinfo{number}{7} (\bibinfo{year}{2011}), \bibinfo{pages}{846--894}.
\newblock


\bibitem[Koch et~al\mbox{.}(2019)]%
        {koch2019automatic}
\bibfield{author}{\bibinfo{person}{Tobias Koch}, \bibinfo{person}{Marco K{\"o}rner}, {and} \bibinfo{person}{Friedrich Fraundorfer}.} \bibinfo{year}{2019}\natexlab{}.
\newblock \showarticletitle{Automatic and semantically-aware 3D UAV flight planning for image-based 3D reconstruction}.
\newblock \bibinfo{journal}{\emph{Remote Sensing}} \bibinfo{volume}{11}, \bibinfo{number}{13} (\bibinfo{year}{2019}), \bibinfo{pages}{1550}.
\newblock


\bibitem[Kuang et~al\mbox{.}(2020)]%
        {kuang2020real}
\bibfield{author}{\bibinfo{person}{Qi Kuang}, \bibinfo{person}{Jinbo Wu}, \bibinfo{person}{Jia Pan}, {and} \bibinfo{person}{Bin Zhou}.} \bibinfo{year}{2020}\natexlab{}.
\newblock \showarticletitle{Real-time UAV path planning for autonomous urban scene reconstruction}. In \bibinfo{booktitle}{\emph{2020 IEEE International Conference on Robotics and Automation (ICRA)}}. IEEE, \bibinfo{pages}{1156--1162}.
\newblock


\bibitem[Lin et~al\mbox{.}(2022)]%
        {2022urbanscene3d}
\bibfield{author}{\bibinfo{person}{Liqiang Lin}, \bibinfo{person}{Yilin Liu}, \bibinfo{person}{Yue Hu}, \bibinfo{person}{Xingguang Yan}, \bibinfo{person}{Ke Xie}, {and} \bibinfo{person}{Hui Huang}.} \bibinfo{year}{2022}\natexlab{}.
\newblock \showarticletitle{Capturing, reconstructing, and simulating: the urbanscene3d dataset}. In \bibinfo{booktitle}{\emph{European Conference on Computer Vision}}. Springer, \bibinfo{pages}{93--109}.
\newblock


\bibitem[Liu et~al\mbox{.}(2021)]%
        {liu2021aerial}
\bibfield{author}{\bibinfo{person}{Yilin Liu}, \bibinfo{person}{Ruiqi Cui}, \bibinfo{person}{Ke Xie}, \bibinfo{person}{Minglun Gong}, {and} \bibinfo{person}{Hui Huang}.} \bibinfo{year}{2021}\natexlab{}.
\newblock \showarticletitle{Aerial path planning for online real-time exploration and offline high-quality reconstruction of large-scale urban scenes}.
\newblock \bibinfo{journal}{\emph{ACM Transactions on Graphics (TOG)}} \bibinfo{volume}{40}, \bibinfo{number}{6} (\bibinfo{year}{2021}), \bibinfo{pages}{1--16}.
\newblock


\bibitem[Liu et~al\mbox{.}(2022)]%
        {liu2022learning}
\bibfield{author}{\bibinfo{person}{Yilin Liu}, \bibinfo{person}{Liqiang Lin}, \bibinfo{person}{Yue Hu}, \bibinfo{person}{Ke Xie}, \bibinfo{person}{Chi-Wing Fu}, \bibinfo{person}{Hao Zhang}, {and} \bibinfo{person}{Hui Huang}.} \bibinfo{year}{2022}\natexlab{}.
\newblock \showarticletitle{Learning reconstructability for drone aerial path planning}.
\newblock \bibinfo{journal}{\emph{ACM Transactions on Graphics (TOG)}} \bibinfo{volume}{41}, \bibinfo{number}{6} (\bibinfo{year}{2022}), \bibinfo{pages}{1--17}.
\newblock


\bibitem[Mauro et~al\mbox{.}(2014)]%
        {mauro2014unified}
\bibfield{author}{\bibinfo{person}{Massimo Mauro}, \bibinfo{person}{Hayko Riemenschneider}, \bibinfo{person}{Alberto Signoroni}, \bibinfo{person}{Riccardo Leonardi}, \bibinfo{person}{Luc Van~Gool}, {et~al\mbox{.}}} \bibinfo{year}{2014}\natexlab{}.
\newblock \showarticletitle{A unified framework for content-aware view selection and planning through view importance}. In \bibinfo{booktitle}{\emph{Proceedings of the 25th British Machine Vision Conference (BMVC 2014)}}. BMVA Press, \bibinfo{pages}{046--01}.
\newblock


\bibitem[Mishchuk et~al\mbox{.}(2017)]%
        {mishchuk2017working}
\bibfield{author}{\bibinfo{person}{Anastasiia Mishchuk}, \bibinfo{person}{Dmytro Mishkin}, \bibinfo{person}{Filip Radenovic}, {and} \bibinfo{person}{Jiri Matas}.} \bibinfo{year}{2017}\natexlab{}.
\newblock \showarticletitle{Working hard to know your neighbor's margins: Local descriptor learning loss}.
\newblock \bibinfo{journal}{\emph{Advances in neural information processing systems}}  \bibinfo{volume}{30} (\bibinfo{year}{2017}).
\newblock


\bibitem[Perron and Furnon({[n.\,d.]})]%
        {ortools}
\bibfield{author}{\bibinfo{person}{Laurent Perron} {and} \bibinfo{person}{Vincent Furnon}.} \bibinfo{year}{[n.\,d.]}\natexlab{}.
\newblock \bibinfo{booktitle}{\emph{OR-Tools}}.
\newblock Google.
\newblock
\urldef\tempurl%
\url{https://developers.google.com/optimization/}
\showURL{%
\tempurl}


\bibitem[Revaud et~al\mbox{.}(2019)]%
        {revaud2019r2d2}
\bibfield{author}{\bibinfo{person}{Jerome Revaud}, \bibinfo{person}{Cesar De~Souza}, \bibinfo{person}{Martin Humenberger}, {and} \bibinfo{person}{Philippe Weinzaepfel}.} \bibinfo{year}{2019}\natexlab{}.
\newblock \showarticletitle{R2d2: Reliable and repeatable detector and descriptor}.
\newblock \bibinfo{journal}{\emph{Advances in neural information processing systems}}  \bibinfo{volume}{32} (\bibinfo{year}{2019}).
\newblock


\bibitem[Roberts et~al\mbox{.}(2017)]%
        {roberts2017submodular}
\bibfield{author}{\bibinfo{person}{Mike Roberts}, \bibinfo{person}{Debadeepta Dey}, \bibinfo{person}{Anh Truong}, \bibinfo{person}{Sudipta Sinha}, \bibinfo{person}{Shital Shah}, \bibinfo{person}{Ashish Kapoor}, \bibinfo{person}{Pat Hanrahan}, {and} \bibinfo{person}{Neel Joshi}.} \bibinfo{year}{2017}\natexlab{}.
\newblock \showarticletitle{Submodular trajectory optimization for aerial 3d scanning}. In \bibinfo{booktitle}{\emph{Proceedings of the IEEE International Conference on Computer Vision}}. \bibinfo{pages}{5324--5333}.
\newblock


\bibitem[Rublee et~al\mbox{.}(2011)]%
        {rublee2011orb}
\bibfield{author}{\bibinfo{person}{Ethan Rublee}, \bibinfo{person}{Vincent Rabaud}, \bibinfo{person}{Kurt Konolige}, {and} \bibinfo{person}{Gary Bradski}.} \bibinfo{year}{2011}\natexlab{}.
\newblock \showarticletitle{ORB: An efficient alternative to SIFT or SURF}. In \bibinfo{booktitle}{\emph{2011 International conference on computer vision}}. Ieee, \bibinfo{pages}{2564--2571}.
\newblock


\bibitem[Schmid et~al\mbox{.}(2020)]%
        {schmid2020efficient}
\bibfield{author}{\bibinfo{person}{Lukas Schmid}, \bibinfo{person}{Michael Pantic}, \bibinfo{person}{Raghav Khanna}, \bibinfo{person}{Lionel Ott}, \bibinfo{person}{Roland Siegwart}, {and} \bibinfo{person}{Juan Nieto}.} \bibinfo{year}{2020}\natexlab{}.
\newblock \showarticletitle{An efficient sampling-based method for online informative path planning in unknown environments}.
\newblock \bibinfo{journal}{\emph{IEEE Robotics and Automation Letters}} \bibinfo{volume}{5}, \bibinfo{number}{2} (\bibinfo{year}{2020}), \bibinfo{pages}{1500--1507}.
\newblock


\bibitem[Selin et~al\mbox{.}(2019)]%
        {selin2019efficient}
\bibfield{author}{\bibinfo{person}{Magnus Selin}, \bibinfo{person}{Mattias Tiger}, \bibinfo{person}{Daniel Duberg}, \bibinfo{person}{Fredrik Heintz}, {and} \bibinfo{person}{Patric Jensfelt}.} \bibinfo{year}{2019}\natexlab{}.
\newblock \showarticletitle{Efficient autonomous exploration planning of large-scale 3-d environments}.
\newblock \bibinfo{journal}{\emph{IEEE Robotics and Automation Letters}} \bibinfo{volume}{4}, \bibinfo{number}{2} (\bibinfo{year}{2019}), \bibinfo{pages}{1699--1706}.
\newblock


\bibitem[Shen et~al\mbox{.}(2024)]%
        {shen2024gim}
\bibfield{author}{\bibinfo{person}{Xuelun Shen}, \bibinfo{person}{Zhipeng Cai}, \bibinfo{person}{Wei Yin}, \bibinfo{person}{Matthias M{\"u}ller}, \bibinfo{person}{Zijun Li}, \bibinfo{person}{Kaixuan Wang}, \bibinfo{person}{Xiaozhi Chen}, {and} \bibinfo{person}{Cheng Wang}.} \bibinfo{year}{2024}\natexlab{}.
\newblock \showarticletitle{GIM: Learning Generalizable Image Matcher From Internet Videos}.
\newblock \bibinfo{journal}{\emph{arXiv preprint arXiv:2402.11095}} (\bibinfo{year}{2024}).
\newblock


\bibitem[Shi et~al\mbox{.}(2023)]%
        {shi2023openWUSU}
\bibfield{author}{\bibinfo{person}{Sunan Shi}, \bibinfo{person}{Yanfei Zhong}, \bibinfo{person}{Yinhe Liu}, \bibinfo{person}{Jue Wang}, \bibinfo{person}{Yuting Wan}, \bibinfo{person}{Ji Zhao}, \bibinfo{person}{Pengyuan Lv}, \bibinfo{person}{Liangpei Zhang}, {and} \bibinfo{person}{Deren Li}.} \bibinfo{year}{2023}\natexlab{}.
\newblock \showarticletitle{Multi-temporal urban semantic understanding based on GF-2 remote sensing imagery: from tri-temporal datasets to multi-task mapping}.
\newblock \bibinfo{journal}{\emph{International Journal of Digital Earth}}  \bibinfo{volume}{16} (\bibinfo{year}{2023}), \bibinfo{pages}{3321--3347}.
\newblock
Issue 1.
\urldef\tempurl%
\url{https://doi.org/10.1080/17538947.2023.2246445}
\showDOI{\tempurl}


\bibitem[Smith et~al\mbox{.}(2018)]%
        {smith2018aerial}
\bibfield{author}{\bibinfo{person}{Neil Smith}, \bibinfo{person}{Nils Moehrle}, \bibinfo{person}{Michael Goesele}, {and} \bibinfo{person}{Wolfgang Heidrich}.} \bibinfo{year}{2018}\natexlab{}.
\newblock \showarticletitle{Aerial path planning for urban scene reconstruction: A continuous optimization method and benchmark}.
\newblock  (\bibinfo{year}{2018}).
\newblock


\bibitem[Tian et~al\mbox{.}(2017)]%
        {tian2017l2}
\bibfield{author}{\bibinfo{person}{Yurun Tian}, \bibinfo{person}{Bin Fan}, {and} \bibinfo{person}{Fuchao Wu}.} \bibinfo{year}{2017}\natexlab{}.
\newblock \showarticletitle{L2-net: Deep learning of discriminative patch descriptor in euclidean space}. In \bibinfo{booktitle}{\emph{Proceedings of the IEEE conference on computer vision and pattern recognition}}. \bibinfo{pages}{661--669}.
\newblock


\bibitem[Tyszkiewicz et~al\mbox{.}(2020)]%
        {tyszkiewicz2020disk}
\bibfield{author}{\bibinfo{person}{Micha{\l} Tyszkiewicz}, \bibinfo{person}{Pascal Fua}, {and} \bibinfo{person}{Eduard Trulls}.} \bibinfo{year}{2020}\natexlab{}.
\newblock \showarticletitle{DISK: Learning local features with policy gradient}.
\newblock \bibinfo{journal}{\emph{Advances in Neural Information Processing Systems}}  \bibinfo{volume}{33} (\bibinfo{year}{2020}), \bibinfo{pages}{14254--14265}.
\newblock


\bibitem[Varghese et~al\mbox{.}(2018)]%
        {varghese2018changenet}
\bibfield{author}{\bibinfo{person}{Ashley Varghese}, \bibinfo{person}{Jayavardhana Gubbi}, \bibinfo{person}{Akshaya Ramaswamy}, {and} \bibinfo{person}{P Balamuralidhar}.} \bibinfo{year}{2018}\natexlab{}.
\newblock \showarticletitle{ChangeNet: A deep learning architecture for visual change detection}. In \bibinfo{booktitle}{\emph{Proceedings of the European conference on computer vision (ECCV) workshops}}. \bibinfo{pages}{0--0}.
\newblock


\bibitem[Wang et~al\mbox{.}(2024)]%
        {wang2024dust3r}
\bibfield{author}{\bibinfo{person}{Shuzhe Wang}, \bibinfo{person}{Vincent Leroy}, \bibinfo{person}{Yohann Cabon}, \bibinfo{person}{Boris Chidlovskii}, {and} \bibinfo{person}{Jerome Revaud}.} \bibinfo{year}{2024}\natexlab{}.
\newblock \showarticletitle{Dust3r: Geometric 3d vision made easy}. In \bibinfo{booktitle}{\emph{Proceedings of the IEEE/CVF Conference on Computer Vision and Pattern Recognition}}. \bibinfo{pages}{20697--20709}.
\newblock


\bibitem[Wu et~al\mbox{.}(2014)]%
        {wu2014quality}
\bibfield{author}{\bibinfo{person}{Shihao Wu}, \bibinfo{person}{Wei Sun}, \bibinfo{person}{Pinxin Long}, \bibinfo{person}{Hui Huang}, \bibinfo{person}{Daniel Cohen-Or}, \bibinfo{person}{Minglun Gong}, \bibinfo{person}{Oliver Deussen}, {and} \bibinfo{person}{Baoquan Chen}.} \bibinfo{year}{2014}\natexlab{}.
\newblock \showarticletitle{Quality-driven poisson-guided autoscanning}.
\newblock  (\bibinfo{year}{2014}).
\newblock


\bibitem[Xu et~al\mbox{.}(2017)]%
        {xu2017autonomous}
\bibfield{author}{\bibinfo{person}{Kai Xu}, \bibinfo{person}{Lintao Zheng}, \bibinfo{person}{Zihao Yan}, \bibinfo{person}{Guohang Yan}, \bibinfo{person}{Eugene Zhang}, \bibinfo{person}{Matthias Niessner}, \bibinfo{person}{Oliver Deussen}, \bibinfo{person}{Daniel Cohen-Or}, {and} \bibinfo{person}{Hui Huang}.} \bibinfo{year}{2017}\natexlab{}.
\newblock \showarticletitle{Autonomous reconstruction of unknown indoor scenes guided by time-varying tensor fields}.
\newblock \bibinfo{journal}{\emph{ACM Transactions on Graphics (TOG)}} \bibinfo{volume}{36}, \bibinfo{number}{6} (\bibinfo{year}{2017}), \bibinfo{pages}{1--15}.
\newblock


\bibitem[Yang et~al\mbox{.}(2023)]%
        {yang2023urbanbis}
\bibfield{author}{\bibinfo{person}{Guoqing Yang}, \bibinfo{person}{Fuyou Xue}, \bibinfo{person}{Qi Zhang}, \bibinfo{person}{Ke Xie}, \bibinfo{person}{Chi-Wing Fu}, {and} \bibinfo{person}{Hui Huang}.} \bibinfo{year}{2023}\natexlab{}.
\newblock \showarticletitle{UrbanBIS: a large-scale benchmark for fine-grained urban building instance segmentation}. In \bibinfo{booktitle}{\emph{ACM SIGGRAPH 2023 Conference Proceedings}}. \bibinfo{pages}{1--11}.
\newblock


\bibitem[Zelinsky et~al\mbox{.}(1993)]%
        {zelinsky1993planning}
\bibfield{author}{\bibinfo{person}{Alexander Zelinsky}, \bibinfo{person}{Ray~A Jarvis}, \bibinfo{person}{JC Byrne}, \bibinfo{person}{Shinichi Yuta}, {et~al\mbox{.}}} \bibinfo{year}{1993}\natexlab{}.
\newblock \showarticletitle{Planning paths of complete coverage of an unstructured environment by a mobile robot}. In \bibinfo{booktitle}{\emph{Proceedings of international conference on advanced robotics}}, Vol.~\bibinfo{volume}{13}. Citeseer, \bibinfo{pages}{533--538}.
\newblock


\bibitem[Zhang et~al\mbox{.}(2021)]%
        {zhang2021continuous}
\bibfield{author}{\bibinfo{person}{Han Zhang}, \bibinfo{person}{Yucong Yao}, \bibinfo{person}{Ke Xie}, \bibinfo{person}{Chi-Wing Fu}, \bibinfo{person}{Hao Zhang}, {and} \bibinfo{person}{Hui Huang}.} \bibinfo{year}{2021}\natexlab{}.
\newblock \showarticletitle{Continuous aerial path planning for 3D urban scene reconstruction.}
\newblock \bibinfo{journal}{\emph{ACM Trans. Graph.}} \bibinfo{volume}{40}, \bibinfo{number}{6} (\bibinfo{year}{2021}), \bibinfo{pages}{225--1}.
\newblock


\bibitem[Zhou et~al\mbox{.}(2020)]%
        {zhou2020offsite}
\bibfield{author}{\bibinfo{person}{Xiaohui Zhou}, \bibinfo{person}{Ke Xie}, \bibinfo{person}{Kai Huang}, \bibinfo{person}{Yilin Liu}, \bibinfo{person}{Yang Zhou}, \bibinfo{person}{Minglun Gong}, {and} \bibinfo{person}{Hui Huang}.} \bibinfo{year}{2020}\natexlab{}.
\newblock \showarticletitle{Offsite aerial path planning for efficient urban scene reconstruction}.
\newblock \bibinfo{journal}{\emph{ACM Transactions on Graphics (TOG)}} \bibinfo{volume}{39}, \bibinfo{number}{6} (\bibinfo{year}{2020}), \bibinfo{pages}{1--16}.
\newblock


\end{thebibliography}

\newpage
\appendix 
\section*{Supplementary Material} 
This section provides additional details and experimental results. References to sections, figures, and tables from the main paper are highlighted in \mainpaper{blue} for clarity.

\section{Prior Probability Statistics}
\label{sec:t1_prob}

Here we show how we derive the \textit{prior probability} used in \mainpaper{Eq.~1} in the \mainpaper{Sec.~3.2} of main paper.
WUSU~\cite{shi2023openWUSU} is a semantic understanding dataset focusing on urban structure and the urbanization process in Wuhan, the central city of the Yangtze River Economic Belt. The dataset spans key development areas, including Jiang'an District and Hongshan District, covering a total geographic area of nearly [geographic area to be completed]. It provides high-resolution, tri-temporal, and multi-spectral satellite images of these districts with $12$ semantic labels, offering unprecedented detail and continuity in representing urban changes. 

We collected images from two time periods: 2015 to 2016 and 2016 to 2018. The number of changed images, broken down by the $12$ labels, is presented in Table.~\ref{tab:change_prob}. 

UrbanBIS~\cite{yang2023urbanbis} serves as a benchmark for large-scale 3D urban understanding, encompassing urban-level semantic segmentation and building-level instance segmentation. It defines $7$ semantic labels: Terrain, Vegetation, Water, Bridge, Vehicle, Boat, and Building. Each label was matched to its corresponding category in WUSU~\cite{shi2023openWUSU} for consistency in analysis, also shown in Table.~\ref{tab:change_prob}.

\subsection{Candidate View Generation Details}

\begin{figure}[h]
    \centering
    \includegraphics[width=\linewidth]{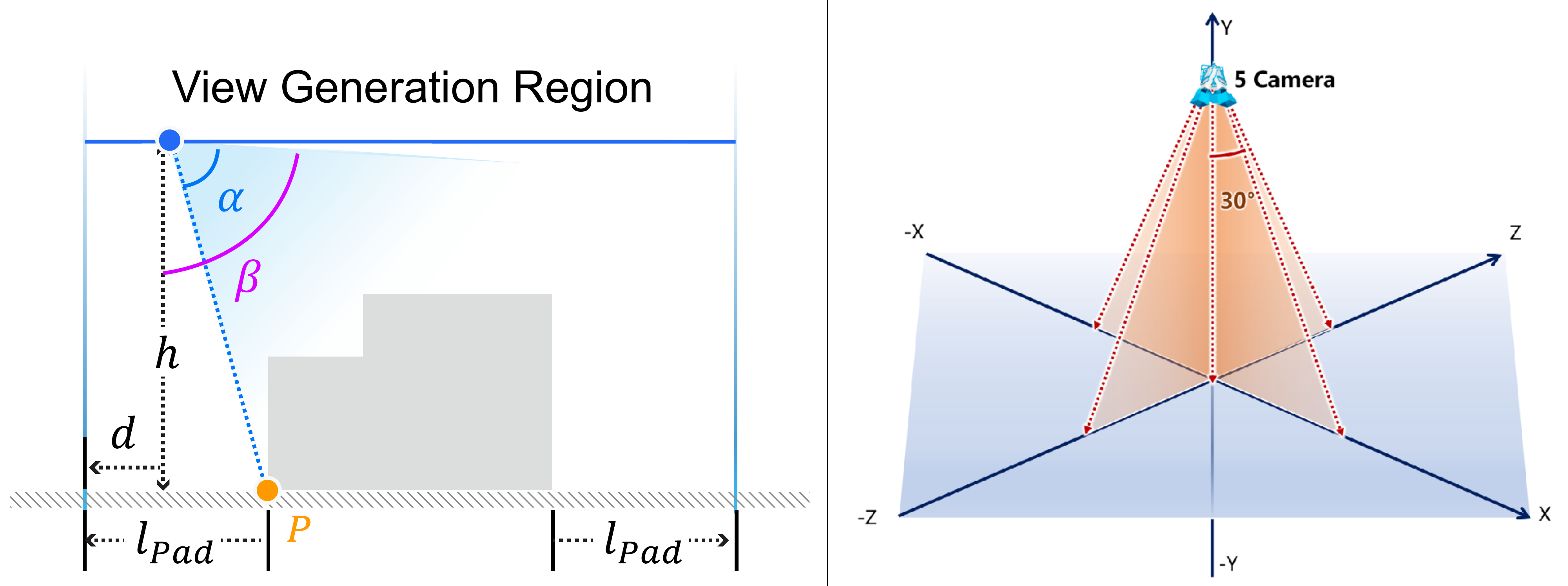}
    \caption{(left) An illustration of our candidate view generation strategy. (right) Five candidate views are generated at the same position, each with a different orientation.}
    \label{fig:padding_camera}
\end{figure}
Given that our primary objective is to detect changes rather than perform a full reconstruction, all candidate viewpoints are generated on a safe-height plane set at $h=120$ meters in our experiments. The region for view generation is determined by expanding the target $\target$ with a padding margin, as illustrated in the inset. 
This padding ensures that the generated views can adequately cover all sample points $\sampleprior_i$ along the edges of $\target$. The padding size $l_{\text{pad}}$ is computed using the camera's field of view (FOV) angles $\alpha$ and $\beta$, along with the safe height $h$, as $l_{\text{pad}} = h \tan(\beta - \alpha) + d$, where $d$ is a small constant to guarantee sufficient coverage. In our setup, $d$ is set equal to the sampling radius used in the subsequent Poisson disk sampling step. The radius parameter, which defines the minimum distance between any two sampled view positions, is set to 15 meters for prior path planning (\mainpaper{Sec.~4}) and 5 meters for real-time path planning (\mainpaper{Sec.~6}). These values are chosen to reflect the different objectives of the two planning stages. In prior path planning, the goal is to achieve broad coverage of the entire scene, which requires a sparser sampling. In contrast, real-time path planning focuses on detailed detection and exploration of localized change areas, thus necessitating a denser sampling.
To accommodate drones equipped with either multiple cameras (e.g., five cameras) or single-camera systems, five candidate views are generated at the same position, each with a different orientation. One view is oriented vertically downward, while the remaining four are tilted at an angle of 30 degrees from the vertical and directed toward the positive x, negative x, positive z, and negative z axes, respectively; see Fig.~\ref{fig:padding_camera} (right) for an illustration.



\section{Prior Path Planning}

\subsection{View optimization objectives}

Recall that the overarching goal for prior path planning is to generate a minimal set of viewpoints $\Viewsprior \subset \Views$ that balances the following competing objectives:  
(1) \textbf{Minimizing Redundancy:} Selecting views such that the total redundancy of the selected subset $\Viewsprior$ is minimized;  
(2) \textbf{Maximizing Coverage:} Guaranteeing that more samples can be observed by selected views in $\Viewsprior$. 
\begin{align}
\Viewsprior = \argmin_{\Viewsprior} \sum_{\viewprior_j \in \Viewsprior} r(\viewprior_j) +
\argmax_{\Viewsprior} \sum_{\sampleprior_i \in \Samplesprior} \cov(\sampleprior_i).
\end{align}
To optimize above two objectives, we use a greedy algorithm that iteratively assesses the importance of each viewpoint and removes the most redundant ones until a stopping criterion is met. The redundancy of a viewpoint $\viewprior_j$ is measured by the redundancy degree described in \mainpaper{Sec.~4} in the main paper. 

The greedy algorithm proceeds as follows: (1) For each viewpoint $\viewprior_j$, compute its redundancy degree $r(\viewprior_j)$; (2) Identify the viewpoint $\viewprior_j$ with the highest redundancy degree. (3) Removing $\viewprior_j$ affects the changeability values of all samples $\sampleprior_i$ observed by $\viewprior_j$. (4) Update the changeability values for these samples to reflect the removal. (5) Repeat.

During each removal step, a hard constraint is enforced to ensure full coverage of the $T_1$ surface. Specifically, if removing a viewpoint $\viewprior_j$ results in any sample $\sampleprior_i$ no longer being observed by any remaining viewpoints, the removal of $\viewprior_j$ is reverted, and the next most redundant viewpoint is considered.  
The process continues iteratively until either all samples $\sampleprior_i \in \Samples$ have at least one observing viewpoint or further removal of viewpoints would violate the coverage constraint or significantly reduce the overall changeability.

\subsection{Trajectory}
\label{sec:tsp}

The result of our optimization process yields a set of viewpoints $\Viewsprior = \{\viewprior_j\}$, which represent the locations and orientations from which the drone needs to capture images. Following a similar approach to \citet{zhou2020offsite}, we construct a continuous trajectory that visits all these viewpoints by formulating the problem as a Traveling Salesman Problem (TSP).

In this formulation, each viewpoint is treated as a node in a fully-connected graph, and the edges represent the cost of traveling between viewpoint pairs. The cost function $\c(\viewprior_i, \viewprior_j)$ for an edge connecting viewpoints $\viewprior_i$ and $\viewprior_j$ is defined as the Euclidean distance $l(\viewprior_i, \viewprior_j)$. The total number of nodes corresponds to the number of remaining viewpoints $|\Viewsprior|$, while the total number of edges is $|\Viewsprior|^2$. 
To solve this TSP efficiently, we utilize the Google OR-Tools package~\cite{ortools}, which provides state-of-the-art algorithms for routing and optimization problems. By solving the TSP for the constructed graph, we obtain an optimal path that minimizes the total travel distance while ensuring all selected viewpoints are visited. This approach guarantees an efficient and continuous flight trajectory tailored to the optimized viewpoints.

\section{Error and Completeness Metrics}
\label{sec:error_completeness}

Following the methodology of \citet{smith2018aerial}, we evaluate the quality of each reconstructed change area using two key metrics: \textit{Error} and \textit{Completeness}, used in \mainpaper{Sec.~7.3} in the main paper.
\begin{itemize}
    \item \textit{Error}: This metric assesses how closely the reconstructed mesh approximates the ground truth in the benchmark dataset. For each point $p_i$ in the reconstructed mesh, the minimum distance to points $\{q_j\}$ in the ground truth is computed as $d_i = |p_i - q_j^{\text{min}}|$. We then analyze the distribution of these distances and determine the 90\% error (and the 95\% error), defined as the smallest distance (in meters) such that 90\% (or 95\%) of the distances $\{d_i\}$ are below this value. A smaller value indicates a higher-quality reconstruction.

    \item \textit{Completeness}: This metric evaluates the coverage of the ground truth by the reconstructed mesh. It is computed in the reverse direction of the \textit{Error} metric. For each point $q_j$ in the ground truth, the minimum distance $d_j$ to points $\{p_i\}$ in the reconstructed mesh is calculated. Given a threshold distance $d$ (in meters), we determine the percentage of ground truth points whose $d_j$ is smaller than $d$. A higher percentage indicates better coverage of the ground truth.
\end{itemize}

\section{Large Scene Statistics}  
\label{sec:large}  

We evaluate our method on two large urban scenes from UrbanBIS~\cite{yang2023urbanbis}, as described in \mainpaper{Sec.~7.4} in the main paper. Table~\ref{tab:large_scene} provides detailed information about these scenes. In this evaluation, we show the number of ground truth change areas (\#GT change) with the number of change areas detected by our method (\#Detected change).  

Due to potential over-segmentation by our method, where a single change area may be incorrectly split into multiple smaller areas, the number of detected change areas can exceed the ground truth count, \eg 18 out of 17 in Yuehai in Table.~\ref{sec:large}. Failure cases resulting from this limitation are further analyzed in Sec.~\ref{sec:failure}.

\section{Real-World Scene Result Analysis}
\label{sec:exp_real_zoomin}

\begin{figure*}
    \centering
    \includegraphics[width=0.8\linewidth]{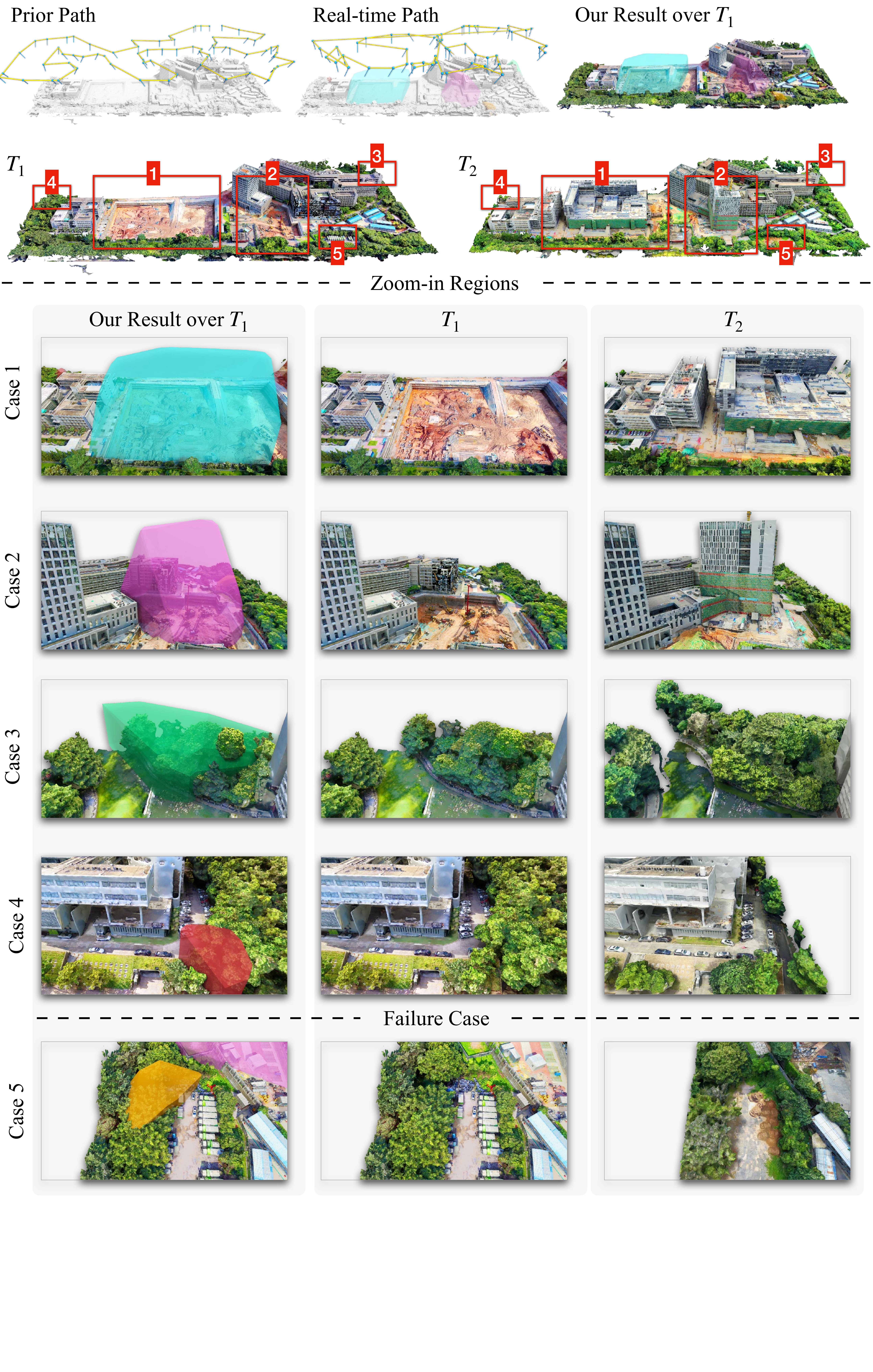}
    \caption{Results on a real-world scene featuring newly constructed buildings and updated areas. Our method successfully detects all change areas, though Case 5 produces an incorrect convex hull. This failure case is analyzed in Sec.~\ref{sec:exp_real_zoomin}. }
    \label{fig:exp_real_zoomin}
\end{figure*}

We evaluated our method on a real-world scene featuring newly constructed buildings and several updated areas, as shown in Fig.~\ref{fig:exp_real_zoomin}. Our method successfully detected five change areas, of which four were correctly identified, while one represented a failure case.

Even in the failure case (Case 5 in Fig.~\ref{fig:exp_real_zoomin}), the change area was correctly detected. However, the convex hull extracted using DUSt3R \cite{wang2024dust3r} was limited due to reliance on only the nearest eight images from the recent view sequence, as described in \mainpaper{Sec.~5}. This limitation highlights a potential area for improvement.
By replacing DUSt3R with a faster and more robust reconstruction algorithm, the efficiency and accuracy of our real-time exploration process could be significantly enhanced.

\begin{table*}[h]
\caption{Statistics of prior probabilities for $12$ semantic labels collected from the WUSU~\cite{shi2023openWUSU} dataset. We also show the semantic label mapping to UrbanBIS~\cite{yang2023urbanbis}. Detail described in Sec.~\ref{sec:t1_prob}}
\begin{center}
\scalebox{1}{
\begin{tabular}{c||c|c|c||c}
\hline
Semantic Label of WUSU &  \#Change & \#Total & Change Prob & Semantic Label of UrbanBIS\\
\hline
Road & 568778 & 31893692 & 0.017833558 & Terrain, Vehicle\\
\hline
Low building  & 2960276  & 39230330  & 0.075458861 & Building $\leq 40$m\\
\hline
High building & 1228858  & 68012494  & 0.018068121 & Building $> 40$m \\
\hline
Arable land & 0 & 317516 & 0 & -\\
\hline
Woodland  & 1748983  & 42439196  & 0.041211502 & Vegetation\\
\hline
Grassland  & 4459227  & 17008223  & 0.262180652 & - \\
\hline
River & 916 & 38543080 & 2.37656E-05 & -\\
\hline
Lake & 94552 & 39519696 & 0.002392529 & Water, Boat\\
\hline
Structure & 978756 & 16369858 & 0.059790134 & Bridge\\
\hline
Excavation & 8293856 & 29785891 & 0.278449149 & -\\
\hline
Bare surface & 32640 & 65788 & 0.496139114 & - \\
\hline
Unclassified & 0 & 261919644 & 0 & - \\
\hline
\end{tabular}}
\end{center}
\label{tab:change_prob}
\end{table*}

\begin{table*}[h]
\caption{Statistics of applying our algorithm on two large scenes, described in Sec.~\ref{sec:large}. Failure cases are shown in Sec.~\ref{sec:failure}}
\begin{center}
\scalebox{1}{
\begin{tabular}{c||c|c|c|c|c|c}
\hline
Scene &  Area ($m^2$) & Height Range (m) &  \#GT change & \#Detected change & \#Views & Path Length (m) \\
\hline
Yuehai &  814,900	& 6-81 & 17	& 18 & 289 & 17,564 \\
\hline
Qingdao	& 405,500	& 5-105 & 22 & 22 & 269 & 16,417 \\
\hline
\end{tabular}}
\end{center}
\label{tab:large_scene}
\end{table*}

\section{Further Failure Case Analysis}
\label{sec:failure}

\begin{figure*}[h]
    \centering
    \includegraphics[width=\linewidth]{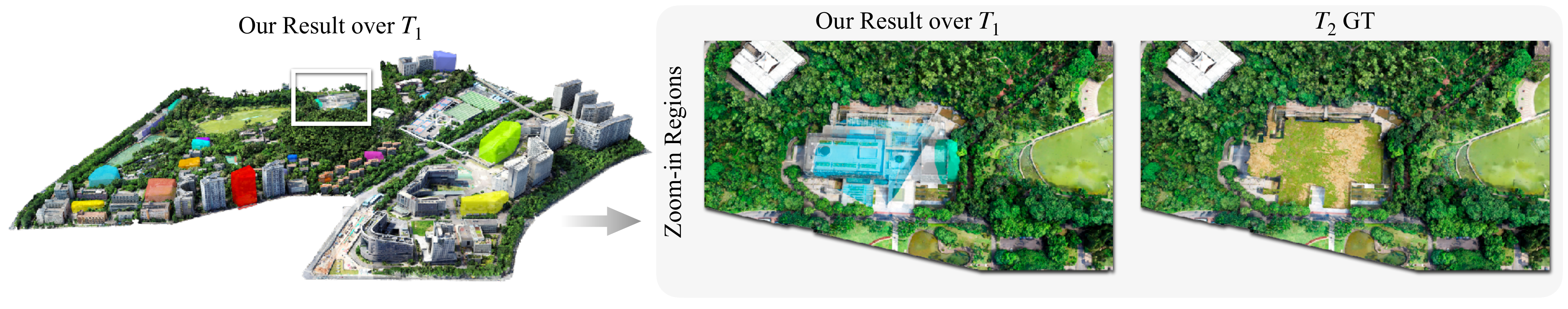}
    \caption{Failure case where the ground truth (GT) change area is a single region, but our method incorrectly identifies it as two separate convex hulls.}
    \label{fig:fig_failure}
\end{figure*}

In addition to the failure case discussed in Sec.\ref{sec:exp_real_zoomin}, we identified another type of failure scenario where a single change area is misidentified as two separate change areas, as illustrated in Fig.\ref{fig:fig_failure}.

The underlying cause of this issue is similar to the one described in Sec.~\ref{sec:exp_real_zoomin} and Fig.~\ref{fig:exp_real_zoomin}. While our method accurately detects the location of the change, the limitations of the DUSt3R~\cite{wang2024dust3r} algorithm become evident. The extraction process relies solely on the nearest eight images from the recent view sequence to maintain online processing speed, which compromises the reconstruction accuracy.

Addressing this limitation with a more precise and efficient reconstruction algorithm could further enhance the reliability of our method in identifying change areas.

\section{Additional Results of Change Area Reconstruction}

We present all three examples of change area reconstruction using our convex hulls as input, compared against the coarse proxy approach from~\citet{zhou2020offsite} in Fig.~\ref{fig:exp_recon}. While the main paper highlights two scenes in~\mainpaper{Fig.~6} with corresponding statistics in~\mainpaper{Table 2}, this section provides the full set of results. These examples demonstrate that our method achieves reconstruction quality comparable to existing approaches while maintaining efficiency and effectiveness.

\begin{table*}[t!]
\small
\caption{Quantitative comparison of three reconstructed scenes using either the \textit{coarse proxy} approach from~\cite{zhou2020offsite} or the convex hulls generated by our method. Visual results are presented in Fig.~\ref{fig:exp_recon}, and the definitions of the two evaluation metrics are provided in Sec.~\ref{sec:error_completeness}.}
\vspace{-10pt}
\begin{center}
\scalebox{1}{
\begin{tabular}{c||c||c|c|c||c|c|c}
\hline
Scene & Input Method & 
\begin{tabular}{@{}c@{}}Error[m] \\  $85\%$ $\downarrow$ \end{tabular} &
\begin{tabular}{@{}c@{}}Error[m] \\ $90\%$ $\downarrow$ \end{tabular} & 
\begin{tabular}{@{}c@{}}Error[m] \\ $95\%$ $\downarrow$ \end{tabular} & 
\begin{tabular}{@{}c@{}}Comp[\%] \\ $0.02$m $\uparrow$ \end{tabular} & 
\begin{tabular}{@{}c@{}}Comp[\%] \\ $0.05$m $\uparrow$ \end{tabular} & 
\begin{tabular}{@{}c@{}}Comp[\%] \\ $0.075$m $\uparrow$ \end{tabular} \\
\hline
\multirow{2}{*}{\hfil Scene 1} 
& Coarse Proxy &	$0.091$ & 	$0.122$ &	$0.197$ & $\mathbf{32.13}$ & $\mathbf{63.09}$ & $\mathbf{76.81}$ \\
& Ours & $\mathbf{0.084}$ & $\mathbf{0.110}$ & $\mathbf{0.171}$ & $29.54$ & $60.86$ &	$75.77$ \\
\hline

\multirow{2}{*}{\hfil Scene 2} 
& Coarse Proxy & $\mathbf{0.081}$ & $\mathbf{0.103}$ & $0.187$ & $31.74$ & $63.47$ & $79.21$ \\
& Ours &	$0.082$ & $0.107$ &	$\mathbf{0.172}$ & $\mathbf{33.68}$ &	$\mathbf{67.76}$ & $\mathbf{80.58}$ \\
\hline

\multirow{2}{*}{\hfil Scene 3} 
& Coarse Proxy & $0.138$ & $0.216$ & $0.396$ &	$\mathbf{24.20}$ & $52.58$ & $\mathbf{67.58}$ \\
& Ours & $\mathbf{0.113}$ & $\mathbf{0.170}$ & $\mathbf{0.312}$ & $24.04$ & $\mathbf{53.48}$ & $67.41$ \\
\hline
\end{tabular}}
\end{center}
\label{tab:exp_recon}
\end{table*}

\begin{figure*}
    \includegraphics[width=0.85\linewidth]{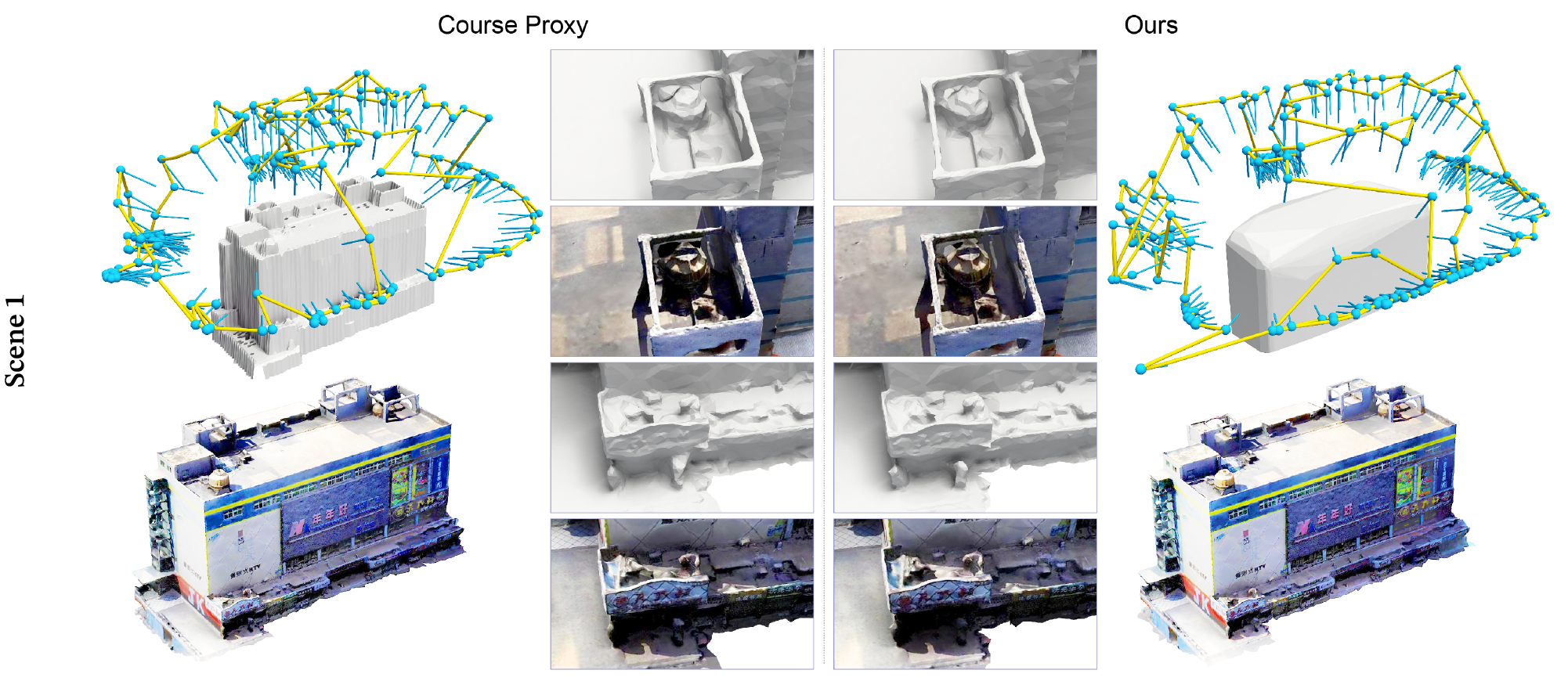} 
    \hfill
    \includegraphics[width=0.85\linewidth]{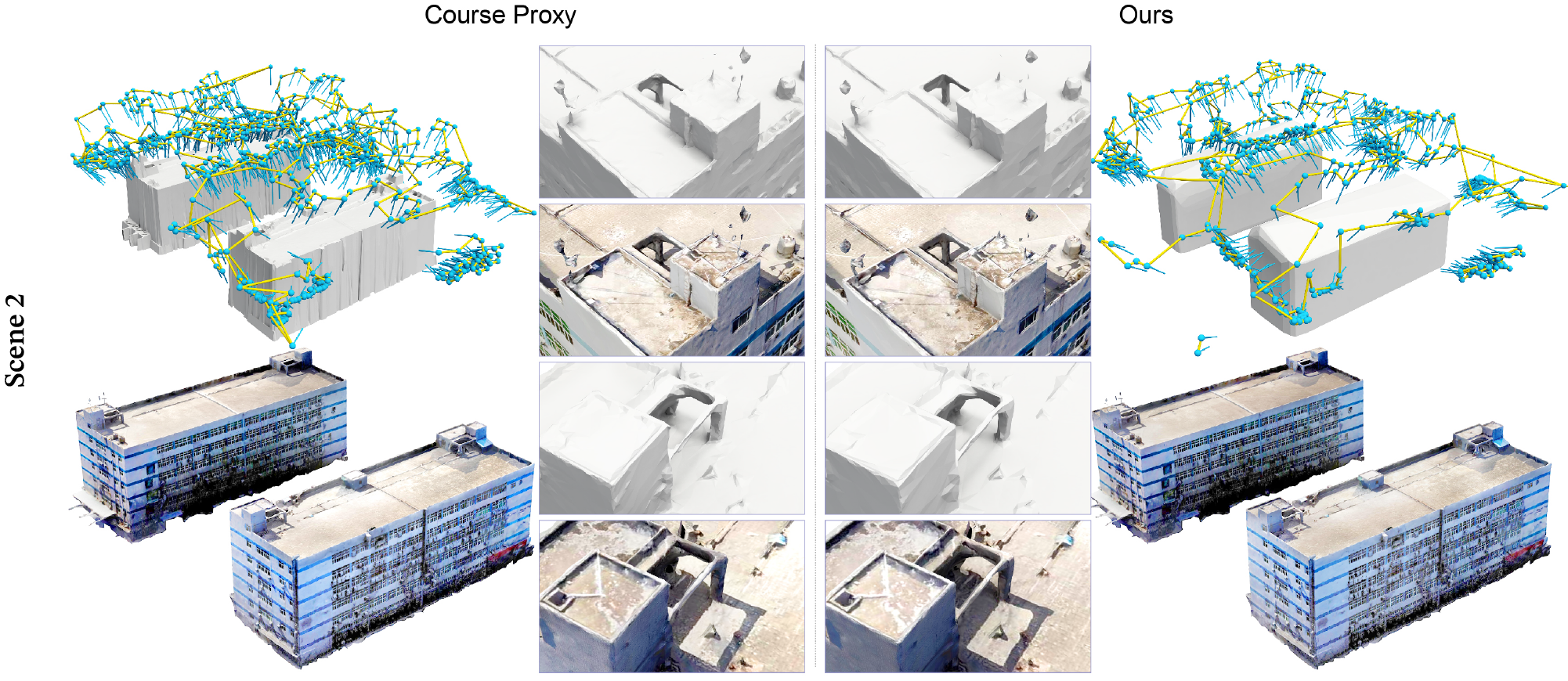} 
    \hfill
    \includegraphics[width=0.85\linewidth]{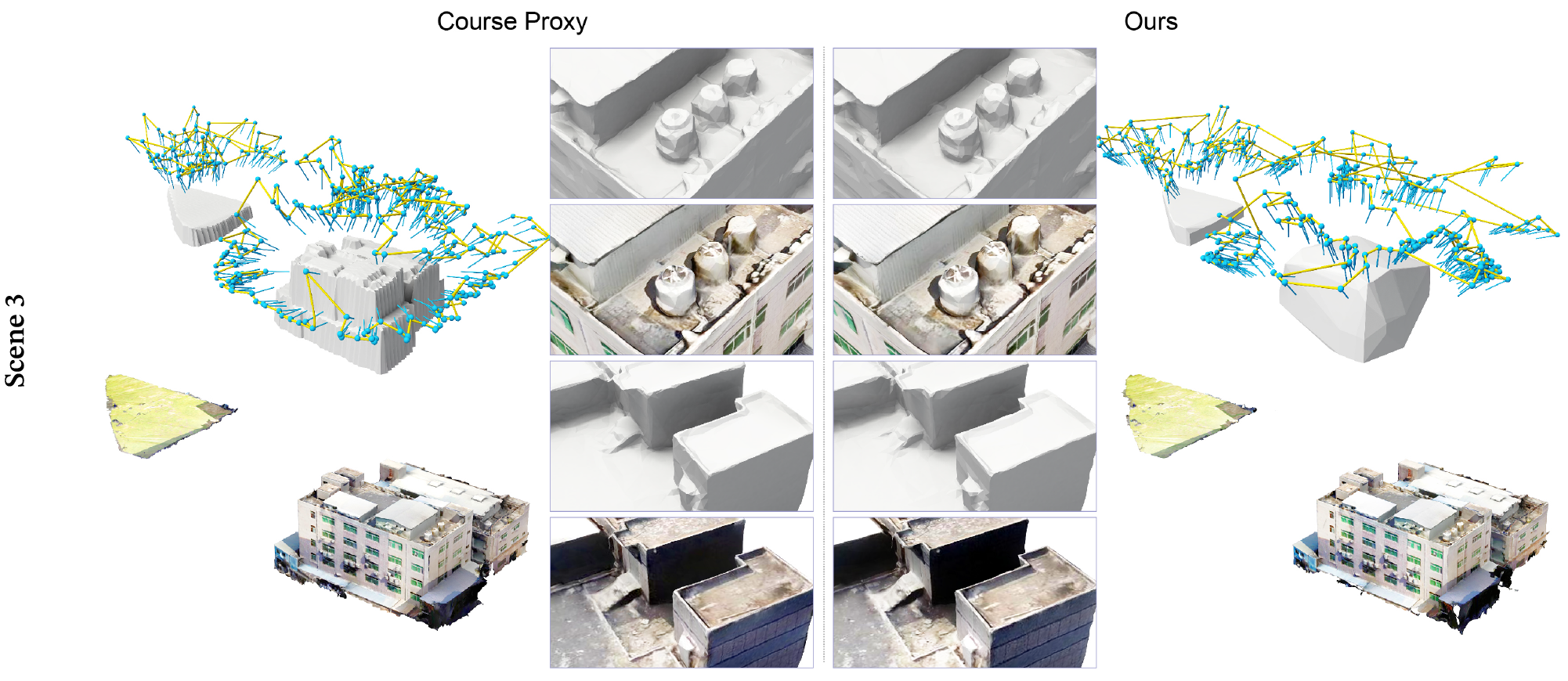} 
    \caption{Visual comparison of reconstructed buildings using either the \textit{coarse proxy} approach from~\cite{zhou2020offsite} or the convex hulls generated by our method. Quantitative statistics corresponding to these results are provided in \mainpaper{Table~2} in the main paper. Reconstructions are shown for three different scenes, demonstrating that our method achieves comparable quality to existing approaches.}
    \label{fig:exp_recon}
\end{figure*}

\begin{algorithm*}
\label{alg:online_planning}
\caption{Change Area Detection and Real-time Path Planning in \mainpaper{Sec.~5 and Sec.~6}}
\SetAlgoLined
\KwIn{prior samples $\Samplesprior$, prior views $\Viewsprior$}
\KwResult{change target set $\target= \{\targetrealtime\}_{t=1}^M$ represents the convex hull of $M$ detected change areas}

\priorviews $\leftarrow$ $\Viewsprior$

\nextview $\leftarrow$ PopOne(\priorviews)

$\pointcloud^t$ target point cloud $\leftarrow$ NULL

\While{\nextview exists}{
    push \nextview to visited set $\Viewsvisited$

    $\image_j^{T_2} \leftarrow$ capture an image using UAV at \nextview at time $T_2$
    
    $\image_j^{T_1} \leftarrow$ get a virtual image at \nextview at $T_1$ in virtual scene
    
    $\diffarea_j \leftarrow$ GIM\_Match($\image_j^{T_1}$, $\image_j^{T_2}$)~\cite{shen2024gim}
    
        
    $\{\pointcloud^t_j\}$ multiple candidate targets 
    $\leftarrow$ DUSt3R($\{\image_{j-8}^{T_2}, \dots, \image_{j}^{T_2}\}$, $\{\diffarea_{j-8}, \dots, \diffarea_{j}\}$)~\cite{wang2024dust3r}
    
    $\pointcloud^t_j \leftarrow$ select nearest change area relative to the drone’s current position from $\{\pointcloud^t_j\}$

    $\prevview \leftarrow \nextview$, $\nextview \leftarrow$ NULL
    
    \If{($\pointcloud^t$ exists) and ($\pointcloud^t_j$ exists) and (IoU($\pointcloud^t_j$, $\pointcloud^t$) $< \phi$)}{
        $\pointcloud^t \leftarrow \pointcloud^t_j \cup \pointcloud^t$
    }
    \If{($\pointcloud^t$ not exists) and ($\pointcloud^t_j$ exists)}{
        $\pointcloud^t \leftarrow \pointcloud^t_j$
    }
    
    \If{$\pointcloud^t$ exist}{
        $\targetrealtime \leftarrow$ compute convex hull from $\pointcloud^t$
        
        \candidateview $\leftarrow$ GenerateTopKCandidateViews($\targetrealtime$, $\Samplesprior$, $\Viewsprior$, $\Viewsvisited$, 10) in Algorithm~\ref{alg:top_K_candidate_view_gen}
        
        \eIf{\candidateview not empty}{
            \nextview $\leftarrow$ select one view nearest to $\prevview$ from \candidateview
        }{
            push $\targetrealtime$ to $\target$

            reset $\pointcloud^t$ $\leftarrow$ NULL
        }
    }
    
    \While{(\nextview not exists) and (\priorviews~not empty)}{
        \prevview $\leftarrow$ select and erase one view nearest to \prevview from \priorviews        
        
        \If{exists a sample in the visual field of $\prevview$ have not been seen by any visited view in $\Viewsvisited$}{        
            \nextview $\leftarrow$ \prevview
            
            break
        }
    }
}
\end{algorithm*}

\begin{algorithm*}
\caption{Top K Candidate Views Generation}
\label{alg:top_K_candidate_view_gen}
\SetAlgoLined
\KwIn{change target $\targetrealtime$, prior samples $\Samplesprior$, prior views $\Viewsprior$, visited views $\Viewsvisited$, parameter K}
\KwOut{top K candidate views set \candidateview}
$\Viewsrealtime \leftarrow$ generate views above the $\targetrealtime$

$\Samplesrealtime \leftarrow$ sample new points on the surface of $\targetrealtime$

$\Samples \leftarrow \Samplesrealtime \bigcup \Samplesprior$

$\Viewsunvisited \leftarrow$ compute all candidate views using \mainpaper{Eq.~(10)}

\If{$\Viewsunvisited$ is empty}{
    return $\emptyset$
} 

compute changeability gain $g(\view_{j+1} | \Viewsvisited)$ for every $\view_{j+1} \in \Viewsunvisited$ using \mainpaper{Eq.~(9)}

\Return the top K views in $\Viewsunvisited$ with the highest $g(\view_{j+1} | \Viewsvisited)$
\end{algorithm*}

\begin{table*}[t!]
\small
\caption{Runtime analysis for all test scenes. References to figures from the main paper are highlighted in \mainpaper{blue} for clarity. The \textit{average real-time path planning time} is the average computation time per frame for selecting the next-best view, excluding the time consumed by replaceable components such as DUSt3R for reconstruction.}
\vspace{-10pt}
\begin{center}
\scalebox{1}{
\begin{tabular}{c||c|c|c|c|c}
\hline
Scene &
\begin{tabular}{@{}c@{}}Prior Path \\ Planning Time [s] \end{tabular} &
\begin{tabular}{@{}c@{}}Avg GIM \& DUSt3R \\ Process Time [s] \end{tabular} &
\begin{tabular}{@{}c@{}} Avg Real-time Path \\ Planning Time [s]\end{tabular} &
\#Views & Figure References \\
\hline
Scene 1 & 0.029 & 10.03 & 0.064 & 9  & \mainpaper{Fig. 6}, Fig. ~\ref{fig:exp_recon} \\
\hline
Scene 2 & 0.047 & 12.25 & 0.091 & 14 & \mainpaper{Fig. 6}, Fig. ~\ref{fig:exp_recon} \\
\hline
Scene 3 & 0.041 & 11.54 & 0.090 & 12 & \mainpaper{Fig. 5}, \mainpaper{Fig. 6}, Fig. ~\ref{fig:exp_recon} \\
\hline
Scene 4  & 0.039 & 11.27 & 0.082 & 35 & \mainpaper{Fig. 4} \\
\hline
yuehai   & 6.727 & 16.27 & 1.225 & 271 & \mainpaper{Fig. 1}, \mainpaper{Fig. 7}, Fig. ~\ref{fig:fig_failure} \\
\hline
qingdao  & 2.493 & 16.23 & 2.252 & 148 & \mainpaper{Fig. 7} \\
\hline
real-world scene & 0.077 & 16.16 & 1.297 & 67 & \mainpaper{Fig. 8}, Fig. ~\ref{fig:exp_real_zoomin} \\
\hline
 \end{tabular}}
\end{center}
\label{tab:runtime_analysis}
\end{table*}


\end{document}